\documentclass{article}
\usepackage{iclr2027_conference,times}

\usepackage{hyperref}
\usepackage{url}

\usepackage{booktabs}
\usepackage{enumitem}
\usepackage{graphicx}
\usepackage{xcolor}
\usepackage{float}
\usepackage{caption}
\usepackage{amsmath,amssymb}
\usepackage{bm}
\usepackage{amsthm}
\usepackage{multirow}
\usepackage{wrapfig}

\theoremstyle{definition}
\newtheorem{theorem}{Theorem}
\newtheorem{lemma}{Lemma}
\newtheorem{corollary}{Corollary}

\usepackage{amsmath,amsfonts,bm}

\def\Figref#1{Figure~\ref{#1}}

\def\Secref#1{Section~\ref{#1}}
\def\eqref#1{equation~\ref{#1}}
\def\Eqref#1{Equation~\ref{#1}}
\def\1{\bm{1}}

\DeclareMathAlphabet{\mathsfit}{\encodingdefault}{\sfdefault}{m}{sl}
\SetMathAlphabet{\mathsfit}{bold}{\encodingdefault}{\sfdefault}{bx}{n}

\newcommand{\R}{\mathbb{R}}

\newcommand{\msg}[1]{\medskip\par\noindent\textbf{#1}\par\smallskip\noindent}

\title{The Reference Matters: Displacement \\ Probing for Weight-Space Learning}

\iclrfinalcopy

\author{\makebox[\dimexpr\textwidth-2\tabcolsep\relax][c]{%
\begin{tabular}[t]{@{}c@{\hspace{3em}}c@{\hspace{3em}}c@{}}
Haechan Shin\thanks{Equal contribution.} & Hoonki Lee\footnotemark[1] & Jungwoon Kang \\
\normalfont Yonsei University & \normalfont Agency for Defense Development & \normalfont Yonsei University \\[1.5ex]
Wooyeol Baek & Sangyeob Kim\thanks{Corresponding authors: \texttt{\{sangyeob.kim, jy.kim\}@yonsei.ac.kr}.} & Jongyoo Kim\footnotemark[2] \\
\normalfont Yonsei University & \normalfont Yonsei University & \normalfont Yonsei University
\end{tabular}}}

\begin{document}

\maketitle
\lhead{Preprint.}

\begin{abstract}
Weight-space learning predicts a model's attributes directly from its parameters, and probing-based methods do so efficiently by reading a model's responses to learned probes. Prior work shows that probing succeeds within a Model Tree—models fine-tuned from a shared pretrained origin—but not why. We trace this to the shared origin itself: since a tree's leaves start from the same weights, what distinguishes them is their displacement from that origin. This motivates probing the displacement rather than the raw weight, an intervention we call the \textbf{DiSplacement Probe (DSProbe)}. We develop this idea along two axes: extending it to second-order Gram responses, whose displacement, to first order, tracks the feature changes induced by fine-tuning; and treating the reference as a design choice, learning a shared component to subtract beyond the pretrained origin. Across four architectures on the Model Jungle benchmark, the resulting \textbf{Refined-DSProbe (R-DSProbe)} sets a new state of the art, matches prior accuracy with substantially fewer probes, and supports a single encoder shared across Model Trees, reducing metanetwork parameters by \(3.61\times\). Code is available at \url{https://github.com/hoonki2002/DSProbe}.
\end{abstract}

\section{Introduction}
\label{sec:intro}

The number of publicly available neural networks is growing rapidly, yet most are released with little documentation of what they were trained on. Weight-space learning offers a way to characterize such models directly from their parameters: it trains metanetworks that take another network's weights as input and predict its attributes---such as its training data or generalization behavior---without external metadata~\citep{eilertsen, unterthiner, schurholt}.

Among these, probing-based methods are the most scalable: rather than processing the full weight tensor, they read a model's responses to a few learned probes, which are compact and permutation-equivariant~\citep{probegen}. ProbeX~\citep{probex} showed that public models form \emph{Model Trees}---checkpoints fine-tuned from a shared ancestor---within which a single layer suffices to predict attributes, and MVProbe~\citep{mvprobe} enriched this with second-order, Gram-based views. Both establish \emph{that} these choices work, but not \emph{why}: what structure in weight space makes probing responses capture the model attributes?

We trace this to the shared pretrained origin of a Model Tree. Because its leaves begin fine-tuning from the same origin, that origin cancels when leaves are compared, so what distinguishes them is their displacement from their reference. This motivates probing the \emph{displacement} rather than the raw weight, an intervention we call the \emph{DiSplacement Probe} (DSProbe). We develop it along two axes. First, the same reasoning extends to the second-order Gram, whose change along training encodes learned features~\citep{nfa, weightgram}; probing the Gram displacement thus records not merely where a weight has moved but what features that movement has learned. Second, we treat the reference itself as a design choice: rather than fixing it to the pretrained origin, we learn a shared component to subtract, isolating the variation that may be more task-specific; we call
this learned-reference extension the \emph{Refined DiSplacement Probe (R-DSProbe)}. All variants keep the architecture and training of prior probing methods unchanged, altering only the probe input.

On the Model Jungle dataset~\citep{probex}, R-DSProbe sets a new state of the art for source-class membership prediction across all four architectures (ResNet~\citep{resnet}, SupViT~\citep{supvit}, MAE~\citep{mae}, and DINO~\citep{dino}).

Our contributions are as follows:
\begin{itemize}
\item We identify the shared pretrained origin as an effective reference that makes leaf-specific variation more distinguishable, and introduce DSProbe to probe this displacement rather than the raw weight.
\item We extend DSProbe to second-order Gram responses, motivate them through feature-learning dynamics, and show that their gain is not explained by expressivity alone.
\item We treat the reference as a design axis rather than a fixed origin, learning a shared component to subtract; this \emph{Refined-DSProbe (R-DSProbe)} reaches state-of-the-art accuracy on Model Jungle across all four architectures.
\end{itemize}

\section{Related Work}
\label{sec:related}

\subsection{Weight-Space Learning}
\label{subsec:wsl_rw}
Early works predict properties such as training configuration~\citep{eilertsen} and generalization behavior~\citep{unterthiner} from the raw parameters of individual models. This line of work has been extended to yield general representations for multiple downstream tasks~\citep{herrmann,schurholt,hyperrep,sane}.

A key challenge with model weights is their high dimensionality and parameter space symmetry~\citep{zhaosymmetry}. Subsequent works incorporate inductive biases for learning from model weights using graph-based representations~\citep{neuralgraphs,gmn, zhang2023neural} and symmetry-aware metanetworks~\citep{dwsnets, nfn, nft}. While symmetry-aware, these approaches still operate directly on parameter space, which limits their scalability to large models.

To address these challenges, probing-based approaches~\citep{probegen,herrmann,probex,mvprobe} have emerged as a dominant paradigm. By characterizing model weights through their responses to learnable probes, these methods naturally avoid weight symmetry, improving scalability to large models. ProbeGen~\citep{probegen} introduces a probe generator to impose structure across learned probes. ProbeX~\citep{probex} observes that public models are organized into \emph{Model Trees}, and improves scalability by extending probing to a single hidden layer within a tree. Most recently, MVProbe~\citep{mvprobe} expands this linear-layer probing with transposed and Gram-based views to recover information missed by the linear response.

While both ProbeX and MVProbe have improved the scalability and expressiveness of single-layer probing, a fundamental question remains unresolved. Our work addresses this gap by showing how shared initialization induces task-relevant model variation and how different probing views expose this variation.

\subsection{Task Vectors}
\label{subsec:task_vec_rw}
Task vectors represent task-specific changes as displacements
between fine-tuned and pretrained model parameters, enabling
model editing through arithmetic operations
\citep{taskarith}.
Subsequent works investigate the effectiveness of task arithmetic
through weight disentanglement
\citep{ortizjimenez2023task}, and improve model merging by
resolving parameter interference, sparsifying task vectors,
or adaptively learning merging coefficients
\citep{yadav2023ties,yu2024language,yang2024adamerging}.
Other studies identify task-specific information within weight
updates or extend arithmetic operations to parameter-efficient
modules \citep{wang2024localizing,zhang2023composing}.
While these approaches primarily use displacements to modify
model behavior, our work probes single-layer displacements
to predict model attributes, further extending this principle
to Gram displacements and learned references.

\subsection{Feature Learning}
\label{subsec:feat_rw}
Weight space reflects training-induced changes in the learned representations.
Feature learning studies how training reshapes the representations learned by a network. Training can align learned features with task-relevant directions~\citep{baratin2021alignment}, while fine-tuning can substantially alter pretrained representations~\citep{kumar}. Such representational changes have been studied through their geometry~\citep{cka,chou2025feature}. The Neural Feature Ansatz~\citep{nfa} relates a layer's weight Gram matrix to the outer-product structure of its feature gradients, with subsequent analyses further characterizing this relationship~\citep{tansley2026nfa}. \citet{weightgram} show that changes in the weight Gram during training track the evolution of feature covariance.

\section{Background}
\label{sec:background}

\subsection{Goal of Weight-Space Learning}
\label{subsec:wsl_bg}
For a task $\tau$ with dataset $D_\tau=\{(x_i,y_i)\}_{i=1}^{n_\tau}$ and target function $f_\tau$, we train $f_\theta$ by minimizing a task-specific loss $\mathcal{L}_\tau$ measuring the mismatch between $f_\theta$ and $f_\tau$:
\begin{equation}
\theta^{(\tau)}
=
\arg\min_{\theta}\mathcal{L}_\tau(\theta),
\qquad
\mathcal{L}_\tau(\theta)
=
\frac{1}{n_\tau}\sum_{i=1}^{n_\tau}
\ell_\tau\left(f_\theta(x_i),f_\tau(x_i)\right).
\end{equation}

Weight-space learning turns this setup around: instead of solving for $\theta^{(\tau)}$, it treats the trained parameters themselves as data. Given a collection of trained models $C=\{(\theta^{(i)},a^{(i)})\}_{i=1}^{N}$, where $a^{(i)}$ is a model attribute of interest (e.g., a source-class label or a CLIP embedding), the goal is to predict the attribute $a^{(i)}$ directly from the parameters $\theta^{(i)}$.

\subsection{Single-Layer Probing within Model Trees}
\label{subsec:slp_bg}
Early work characterizes an entire model through its responses $f_\theta(\mathbf{U})$ to a learnable probe matrix $\mathbf{U}$~\citep{probegen}. Recently, ProbeX~\citep{probex} extends probing to a single hidden layer.
Let $\theta=\{W_l\}_{l=1}^L$ denote the layer weight matrices of a model and let $W_l\in\R^{m\times n}$ be the selected layer, where $n$ and $m$ are input and output dimensions. The probing network for $W_l$ is
\begin{equation}
\Phi_l = W_l \mathbf{U} \in \R^{m\times r},
\qquad
\mathbf{U} = [\mathbf{u}_1,\dots,\mathbf{u}_r]\in\R^{n\times r},
\qquad
\hat a^{(i)}=g(\Phi_l),
\end{equation}
where $\Phi_l$ denotes the probe response of  $W_l$ and $g$ denotes an encoder mapping these responses to the predicted attribute, with probes and encoder trained jointly.

\subsection{Multi-View Probing of a Single Layer}
\label{subsec:mvp_bg}
With a fixed probe matrix $\mathbf{U}$ of rank $r<n$, the linear response $W_l\mathbf{U}$ does not uniquely characterize $W_l$. MVProbe~\citep{mvprobe} targets the limited expressivity of first-order probing by introducing second-order Gram views. The first-order row and column views capture directional responses, while their Gram counterparts capture pairwise relationships, together providing four complementary views:
\begin{equation}
\resizebox{\textwidth}{!}{$\displaystyle
\Phi_{\mathrm{row}} = W_l \mathbf{U}_{\mathrm{row}},\quad
\Phi_{\mathrm{col}} = W_l^\top \mathbf{U}_{\mathrm{col}},\quad
\Phi_{\mathrm{row\text{-}ker}} = W_l W_l^\top \mathbf{U}_{\mathrm{row\text{-}ker}},\quad
\Phi_{\mathrm{col\text{-}ker}} = W_l^\top W_l \mathbf{U}_{\mathrm{col\text{-}ker}}
$}.
\nonumber
\end{equation}

\subsection{Task Vectors as Parameter Displacements}
\label{subsec:task_vec}
Task vectors represent task-specific parameter changes induced
by fine-tuning relative to a pretrained model~\citep{taskarith,ortizjimenez2023task}.
Given pretrained parameters $\theta_{\mathrm{pre}}$ and
parameters $\theta^{(\tau)}$ fine-tuned on task $\tau$,
the corresponding task vector is defined as
\begin{equation}
    \Delta\theta^{(\tau)}
    = \theta^{(\tau)} - \theta_{\mathrm{pre}}.
\end{equation}
This displacement captures the parameter changes induced
by task-specific fine-tuning, excluding the shared pretrained
component. Prior work shows that adding or negating task vectors
can steer model behavior toward or away from the corresponding
tasks~\citep{taskarith}.
These findings suggest that task vectors encode
task-relevant directions in weight space.

\section{Motivation}
\label{sec:questions}

\msg{What makes within-tree variation easier to probe?}
ProbeX~\citep{probex} observes that public models can be organized into Model Trees. A Model Tree has a shared pretrained checkpoint as its \emph{root} and, as its \emph{leaves}, the models obtained by fine-tuning that root on different source tasks. ProbeX shows that within a tree, a single hidden layer can suffice to predict model attributes. It attributes this improvement to reduced \emph{nuisance variation}, i.e., weight variation unrelated to the target attribute, but does not explain why model attributes are accessible from the remaining variation.
Following the task-vector perspective~\citep{taskarith},
we characterize this variation through layer-wise displacements
from the shared pretrained origin in \Secref{subsec:shared_origin}.

\msg{Does greater expressivity guarantee better probing?}
MVProbe~\citep{mvprobe} motivates second-order Gram views
by showing that distinct weight matrices can yield identical
first-order probe responses while remaining distinguishable
under second-order responses~\citep[Thm.~4.1]{mvprobe}.
This separation property extends to higher orders under
suitable dimensional conditions
(Appendix~\ref{app:higher_order},
Theorem~\ref{thm:higher_order_app}).
However, greater separation capacity does not necessarily
imply better attribute prediction.

\begin{wraptable}{r}{0.49\textwidth}
\centering
\caption{
Model Jungle accuracy (\%) with higher-order MVProbe
branches and capacity-matched controls.
}
\label{tab:higher_order}
\scriptsize
\setlength{\tabcolsep}{2.4pt}
\renewcommand{\arraystretch}{1.08}
\begin{tabular}{lcccc}
\toprule
Method & ResNet & SupViT & MAE & DINO \\
\midrule
MVProbe
    & 91.09 & 92.41 & 81.24 & 77.88 \\
Degree-3
    & \textbf{93.66} & 92.78 & 81.12 & 77.18 \\
Capacity control
    & 92.36 & 92.75 & 81.49 & 78.63 \\
Degree-3+4
    & 93.46 & 92.83 & 81.36 & 77.04 \\
Capacity control
    & 92.22 & \textbf{93.24} & \textbf{82.02}
    & \textbf{78.72} \\
\bottomrule
\end{tabular}
\end{wraptable}

As shown in Table~\ref{tab:higher_order}, augmenting MVProbe
with third- and fourth-order branches does not consistently
improve accuracy over capacity-matched controls.
MVProbe's own higher-order ablation likewise reports
architecture-dependent gains~\citep[Table~9]{mvprobe}.
These findings suggest that expressivity alone does not
fully explain the predictive utility of Gram-based probing.

A complementary perspective comes from feature-learning
dynamics.
For a layer $z=Wh$, \citet{weightgram} define the virtual feature update
$h^{+}=h-\eta\nabla_h\mathcal{L}$ and the resulting
\emph{virtual covariance shift} (VCS): $h^{+}(h^{+})^\top-hh^\top$.
They show that removing the Gram-preserving
component of the weight gradient largely preserves
and sometimes improves training performance, indicating that Gram-changing updates capture a substantial part of the feature-learning signal.
Motivated by this connection, we extend the task-vector
perspective to Gram matrices, probing their changes
relative to the shared pretrained origin
(\Secref{subsec:gram_displacement}).

\section{Method}
\label{sec:method}

\subsection{DiSplacement Probe: Seeing the tree by standing at its root}
\label{subsec:shared_origin}

Within a Model Tree, all leaf models are fine-tuned from a shared
pretrained checkpoint $\theta_{\mathrm{pre}}$ on different source
tasks~\citep{probex}. Let $W_l^{(\tau)}\in\R^{m\times n}$
denote the selected layer of a model fine-tuned on task $\tau$,
and let $W_{\mathrm{pre}}$ be its pretrained counterpart.
Following the task-vector perspective~\citep{taskarith,taskvecgrad},
we express each leaf weight as
\begin{equation}
    W_l^{(\tau)} = W_{\mathrm{pre}} + \Delta_W^{(\tau)},
    \label{eq:weight_displacement}
\end{equation}
where $\Delta_W^{(\tau)}$ is the layer-wise component of the
task vector, corresponding to the accumulated parameter updates
during fine-tuning. Since all leaves share the same initial parameter state, each task induces a distinct gradient direction in the shared parameter space (Appendix~\ref{app:task_vec_update}). When comparing leaves fine-tuned on tasks $\tau$ and $\tau'$,
the shared pretrained component cancels, leaving
$\Delta_W^{(\tau)}-\Delta_W^{(\tau')}$ as their difference.
Motivated by this task-conditioned variation, we probe
$\Delta_W$ through the first-order responses
\begin{equation}
    \Phi_W = \Delta_W \mathbf U_W,
    \qquad
    \Phi_{W^\top} = \Delta_W^\top \mathbf U_{W^\top},
    \label{eq:first_order_displacement_views}
\end{equation}
where $\mathbf U_W\in\R^{n\times r}$ and
$\mathbf U_{W^\top}\in\R^{m\times r}$ are learnable probe matrices.
We call this intervention the \emph{DiSplacement Probe}
(DSProbe); it retains the architecture of single-layer
probing methods, while changing the input to the
\emph{displacement} from its raw counterpart.

\subsection{DiSplacement Probe for Weight Gram}
\label{subsec:gram_displacement}
Analogously to the layer-wise task vector, changes in the
weight Gram accumulate along the fine-tuning trajectory.
For the selected layer at step $t$, let $H_t$ denote the input features induced by task $\tau$ and $Z_t=W_{l,t}H_t$ denote the corresponding output features. Following the analysis of~\citet{weightgram}, we derive that, to the first order in $\eta$, each Gram update corresponds to the virtual covariance shift in the input and output feature space, respectively (Appendix~\ref{app:gram_feature_update}):
\begin{equation}
\begin{aligned}
    W_{l,t+1}^\top W_{l,t+1} - W_{l,t}^\top W_{l,t}
    &=
    -2\eta\,\operatorname{Sym}\!\left(
        \nabla_{H_t}\mathcal{L}_\tau(\theta_t)\,H_t^\top
    \right)
    + \mathcal{O}(\eta^2),
    \\
    W_{l,t+1} W_{l,t+1}^\top - W_{l,t} W_{l,t}^\top
    &=
    -2\eta\,\operatorname{Sym}\!\left(
        \nabla_{Z_t}\mathcal{L}_\tau(\theta_t)\,Z_t^\top
    \right)
    + \mathcal{O}(\eta^2),
\end{aligned}
\label{eq:gram_feature_updates}
\end{equation}
where $\operatorname{Sym}(A):=(A+A^\top)/2$.
Accumulating these task-conditioned Gram updates along the
fine-tuning trajectory yields the endpoint displacements
from the shared pretrained origin:
\begin{equation}
    \Delta_H
    = W_l^\top W_l - W_{\mathrm{pre}}^\top W_{\mathrm{pre}},
    \qquad
    \Delta_Z
    = W_lW_l^\top - W_{\mathrm{pre}}W_{\mathrm{pre}}^\top,
    \label{eq:gram_displacement}
\end{equation}
recording what features that movement has learned for each task.
Together with the first-order displacements $\Delta_W$ and $\Delta_W^\top$, which record where the weight has moved, they yield the following displacement-based response formulations for DSProbe:
\begin{equation}
    \Phi_W = \Delta_W \mathbf U_W,
    \qquad
    \Phi_{W^\top} = \Delta_W^\top \mathbf U_{W^\top},
    \qquad
    \Phi_H = \Delta_H \mathbf U_H,
    \qquad
    \Phi_Z = \Delta_Z \mathbf U_Z.
    \label{eq:displacement_views}
\end{equation}

\subsection{Refined DiSplacement Probe}
\label{subsec:refined}

The gains from probing displacements suggest that removing the components shared across the leaves makes task-relevant variation more accessible. Still, removing the pretrained root may leave such a component, for example the mean displacement of the leaves. We therefore introduce a learnable
correction $C\in\R^{m\times n}$ to account for variation
that may remain shared across leaf displacements.
Writing $W_*=W_{\mathrm{pre}}+C$, the original displacement (\eqref{eq:weight_displacement})
decomposes as $\Delta_W^{(\tau)}
=\widetilde{\Delta}_W^{(\tau)}+C$,
yielding the following refined displacements:
\begin{equation}
\begin{aligned}
    \widetilde{\Delta}_W
    &= W_l-W_*,\qquad
    \widetilde{\Delta}_H
    = W_l^\top W_l-W_*^\top W_*,\qquad
    \widetilde{\Delta}_Z
    = W_lW_l^\top-W_*W_*^\top.
\end{aligned}
\label{eq:refined_displacement}
\end{equation}
We probe these refined displacements using the same
first- and second-order responses as DSProbe
(\eqref{eq:displacement_views}), and call this extension the \emph{Refined DiSplacement
Probe} (R-DSProbe):
\begin{equation}
    \widetilde\Phi_W = \widetilde\Delta_W \mathbf U_W,
    \qquad
    \widetilde\Phi_{W^\top} = \widetilde\Delta_W^\top \mathbf U_{W^\top},
    \qquad
    \widetilde\Phi_H = \widetilde\Delta_H \mathbf U_H,
    \qquad
    \widetilde\Phi_Z = \widetilde\Delta_Z \mathbf U_Z.
    \label{eq:refined_displacement_views}
\end{equation}

In practice, $W_*$ is optimized directly rather than
parameterizing $C$, and initialized with the mean of the
training leaf weights within each tree. This initialization
corresponds to mean centering, while joint optimization
with the probing network adapts the reference to the
prediction objective (\Secref{sec:reference}).
Importantly, this procedure does not require access to
$W_{\mathrm{pre}}$. When tree membership is also unknown,
models can be clustered using weight-space distances~\citep{probex}, and the resulting cluster means provide
the initial references.

\section{Experiments and Results}
\label{sec:experiments}
We follow the Model Jungle task settings of MVProbe~\citep{mvprobe} and evaluate
two settings: \emph{discriminative}, comprising source-class membership prediction
for ResNet~\citep{resnet}, SupViT~\citep{supvit}, MAE~\citep{mae}, and
DINO~\citep{dino}; and \emph{generative}, comprising training-class prediction
for Stable Diffusion~\citep{ldm} LoRAs~\citep{LoRA}. We compare DSProbe and R-DSProbe with StatNN~\citep{unterthiner}, ProbeX~\citep{probex}, and MVProbe, matching probe counts, per-branch encoder dimensions, and encoder output dimensions across probing methods. Each method's best-performing layer is selected based on validation accuracy (\Figref{fig:layerwise}). For SD-LoRA, the probed matrix $BA$ is itself the displacement from the pretrained weight, so DSProbe coincides with MVProbe and we report only R-DSProbe. Dataset specifications, baseline descriptions, and implementation details are provided in Appendix~\ref{app:experiment_details}.

\subsection{Main Results}
Table~\ref{tab:main} summarizes the performance of displacement-based probing on
Model Jungle. In the discriminative setting, DSProbe already substantially improves
over probing the original weights. This supports our central premise that removing
the shared pretrained origin exposes variation that is more discriminative of the
source task.

Refining the reference further strengthens this effect. By learning $W_*$ and
removing task-common variation that remains after pretrained subtraction, R-DSProbe
yields the best performance across all four architectures. This holds at the level
of individual responses as well: evaluating each response in isolation with a
single-branch probe (Appendix~\ref{app:single_branch}), the pretrained displacement
improves 10 of the 12 model--response combinations and the refined displacement
improves all 12. The two exceptions are the second-order responses of DINO, which
pretrained subtraction alone degrades but refinement recovers, explaining the
marginal gain of DSProbe over MVProbe on DINO. The consistent improvement from the
fixed pretrained reference to the learned reference suggests that the choice of
reference itself can be optimized for weight-space prediction.

The same displacement principle remains effective in the generative setting. Here,
in-distribution (ID) evaluation uses test models from classes seen during
training, whereas held-out-class evaluation uses test models from classes not used
for fitting. R-DSProbe attains the highest accuracy on both SD-200 and SD-1k; paired
t-tests over five matched seeds show that its improvement over MVProbe is
statistically significant on SD-1k, whereas on SD-200, where MVProbe is already
strong, the two are comparable (Appendix~\ref{app:significance}). The strong held-out-class
accuracy on SD-1k suggests that the representation retains source-dependent
information beyond the classes observed when fitting the probe
(Appendix~\ref{app:experiment_details} describes how checkpoints are selected). Since the model
architecture and probing capacity are identical across methods, the differences in
performance stem from the input representation.

\begin{table}[t]
\caption{Main results on Model Jungle. Left: source-class membership accuracy (\%)
on discriminative models. Right: classification accuracy (\%) on Stable Diffusion
LoRA models. ProbeX ($\times 4$) matches the parameter budget of the four-branch
methods. Mean over five seeds; standard deviations are reported in
Tables~\ref{tab:main_discriminative} and~\ref{tab:main_generative}.}
\label{tab:main}
\centering

\begin{minipage}[t]{0.52\linewidth}
\centering
\resizebox{\linewidth}{!}{%
\begin{tabular}{lcccc}
\toprule
\textbf{Method} & \textbf{ResNet} & \textbf{SupViT} & \textbf{MAE} & \textbf{DINO}\\
\midrule
StatNN            & $65.83$ & $52.43$ & $50.54$ & $50.53$\\
ProbeX ($\times 4$) & $89.89$ & $89.17$ & $77.02$ & $69.87$\\
MVProbe           & $91.09$ & $92.41$ & $81.24$ & $77.88$\\
\textbf{DSProbe (ours)}   & $96.82$ & $95.93$ & $83.84$ & $77.89$\\
\textbf{R-DSProbe (ours)} & $\mathbf{97.33}$ & $\mathbf{96.51}$ & $\mathbf{85.24}$ & $\mathbf{79.11}$\\
\bottomrule
\end{tabular}%
}
\end{minipage}
\hfill
\begin{minipage}[t]{0.46\linewidth}
\centering
\resizebox{\linewidth}{!}{%
\begin{tabular}{lcccc}
\toprule
& \multicolumn{2}{c}{\textbf{SD-200}} & \multicolumn{2}{c}{\textbf{SD-1k}}\\
\cmidrule(lr){2-3}\cmidrule(lr){4-5}
\textbf{Method} & \textbf{ID} & \textbf{Held-out} & \textbf{ID} & \textbf{Held-out}\\
\midrule
ProbeX ($\times 4$) & $97.92$ & $93.25$ & $32.22$ & $52.77$\\
MVProbe           & $99.80$ & $95.25$ & $98.00$ & $97.09$\\
\textbf{R-DSProbe (ours)} & $\mathbf{99.84}$ & $\mathbf{95.53}$ & $\mathbf{99.04}$ & $\mathbf{98.48}$\\
\bottomrule
\end{tabular}%
}
\end{minipage}
\end{table}

\subsection{Layer-wise Comparison}
\begin{figure}[t]
\centering
\includegraphics[width=\textwidth]{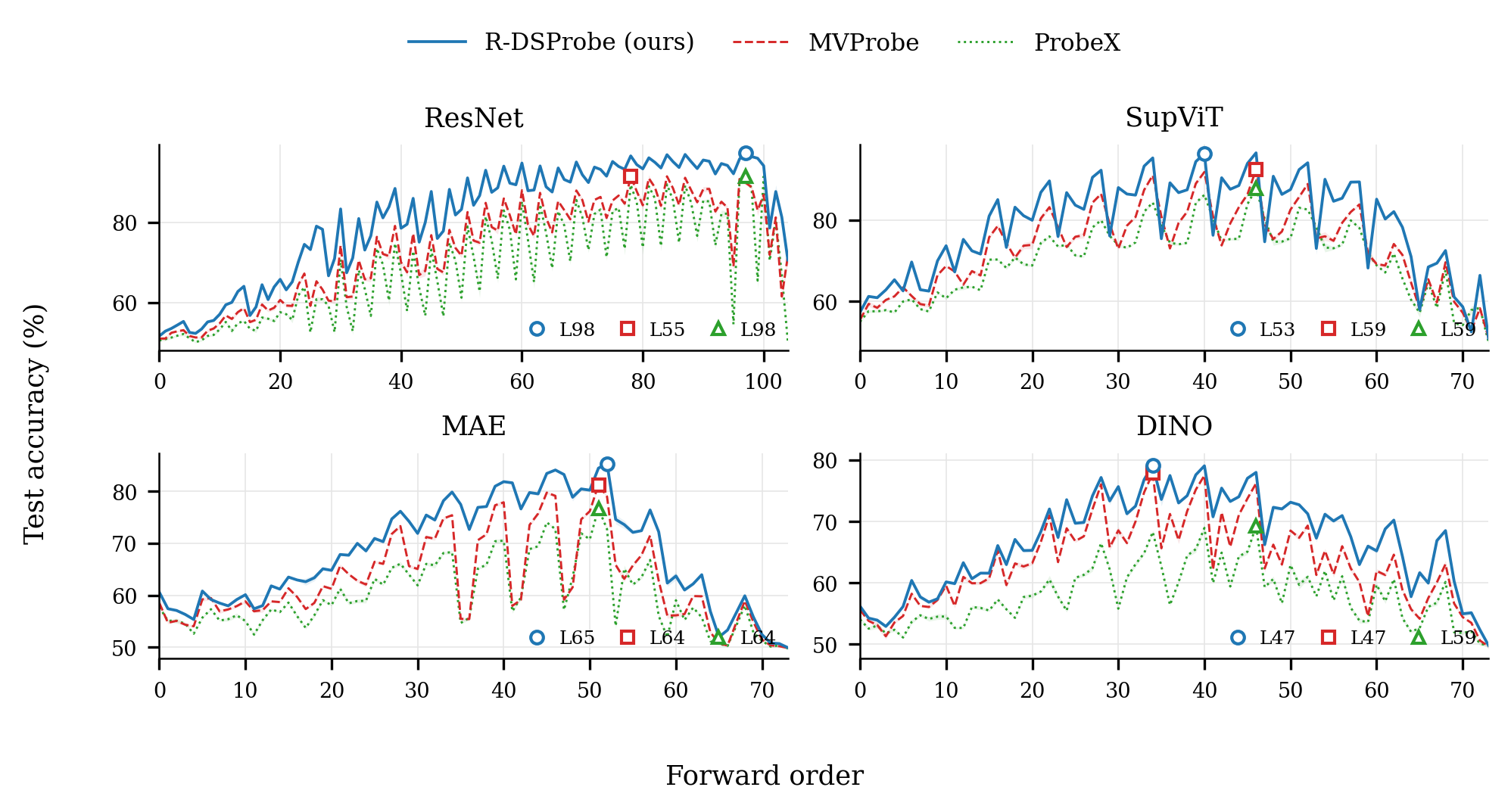}
\caption{Full-layer test accuracy (\%) in forward order for R-DSProbe (blue), MVProbe
(red), and ProbeX (green), averaged over five seeds; open markers denote each
method's validation-selected layer (indices below each panel).}
\label{fig:layerwise}
\end{figure}

In \Figref{fig:layerwise}, we compare the layer-wise performance of R-DSProbe against ProbeX and MVProbe. R-DSProbe is not confined to the fixed layer used in the main experiments. Across all four architectures, R-DSProbe generally tracks the same layer-wise performance profile as ProbeX and MVProbe, but shifts the accuracy curve upward over the most informative layers. The improvement is particularly pronounced for ResNet, where R-DSProbe maintains a substantial margin throughout the deeper layers, and remains visible for SupViT, MAE, and DINO despite their markedly different layer-wise profiles.

Importantly, the layer-wise performance is broadly aligned across methods. Layers that are informative for ProbeX or MVProbe also tend to be informative for R-DSProbe, while displacement probing extracts consistently stronger predictive signals from them. This suggests that R-DSProbe improves the accessibility of task-relevant variation within a layer. Prior work shows that ViTs maintain more uniform representations across depth than ResNets~\citep{raghu2021vitcnn}, and that the target becomes increasingly linearly predictable in deeper layers~\citep{weightgram}.
The resemblance of our layer-wise patterns to these established trends suggests that single-layer probing may offer a useful lens for identifying where task-relevant features occur across depth.

\subsection{The Reference Matters: Comparison with Centering}
\label{sec:reference}

A central choice in DSProbe is to probe the displacement from a reference rather than the raw weight, which raises a natural question: does the gain merely reflect centering the input, or does \emph{what} is subtracted matter? We compare five references under the same four-branch architecture (Table~\ref{tab:reference}): the raw weight ($W$), a random matrix ($W_{\mathrm{rand}}$) rescaled to the Frobenius norm of $W_{\mathrm{pre}}$, the pretrained origin ($W_{\mathrm{pre}}$), the mean of the leaf weights ($W_{\mathrm{mean}}$, i.e., centering), and the learned reference ($W_*$). What is subtracted is decisive. Subtracting a norm-matched $W_{\mathrm{rand}}$ changes accuracy only marginally and in either direction across architectures, whereas referencing the pretrained origin $W_{\mathrm{pre}}$ improves it substantially. Most of the improvement over MVProbe comes from this step: averaged over the four architectures, DSProbe accounts for about three quarters of the gain of R-DSProbe. This also explains why centering helps: since each leaf is $W_l = W_{\mathrm{pre}} + \Delta_W$, subtracting $W_{\mathrm{mean}}$ removes the pretrained component together with the mean displacement, making centering itself a displacement-based representation rather than an alternative to it. The learned reference $W_*$ improves further and attains the highest accuracy on every architecture.

\begin{table}[t]
\caption{Effect of the reference subtracted before probing, under the four-branch
architecture. $W_{\mathrm{rand}}$ is rescaled to the Frobenius norm of
$W_{\mathrm{pre}}$. Accuracy (\%), mean $\pm$ population standard deviation over
five seeds.}
\label{tab:reference}
\centering
\begin{tabular}{lccccc}
\toprule
\textbf{Model}
& \textbf{MVProbe}
& \textbf{Random}
& \textbf{DSProbe}
& \textbf{Centering}
& \textbf{R-DSProbe}\\
& \textbf{($W$)}
& \textbf{($W\!-\!W_{\mathrm{rand}}$)}
& \textbf{($W\!-\!W_{\mathrm{pre}}$)}
& \textbf{($W\!-\!W_{\mathrm{mean}}$)}
& \textbf{($W\!-\!W_*$)}\\
\midrule
ResNet & $91.09 \pm 0.33$ & $91.19 \pm 0.65$ & $96.82 \pm 0.14$ & $95.36 \pm 0.12$ & $\mathbf{97.33 \pm 0.05}$\\
SupViT & $92.41 \pm 0.40$ & $92.28 \pm 0.12$ & $95.93 \pm 0.15$ & $95.47 \pm 0.22$ & $\mathbf{96.51 \pm 0.19}$\\
MAE    & $81.24 \pm 0.20$ & $82.46 \pm 0.19$ & $83.84 \pm 0.15$ & $84.80 \pm 0.21$ & $\mathbf{85.24 \pm 0.27}$\\
DINO   & $77.88 \pm 0.14$ & $76.82 \pm 0.56$ & $77.89 \pm 0.25$ & $77.84 \pm 0.30$ & $\mathbf{79.11 \pm 0.13}$\\
\bottomrule
\end{tabular}
\end{table}

A related concern is that, because the probe is linear, subtracting a reference from the weight is equivalent to adding a constant offset to the response, so the gain might come from the added capacity of a learned offset rather than from what is subtracted. We test this by replacing $W_*$ with a free per-branch offset added to the response. Initialized at zero, it recovers none of the gain ($91.11$ vs.\ $91.09$ for MVProbe on ResNet); initialized at the centering point, it remains below R-DSProbe on every architecture (Appendix~\ref{app:offset}). The gain therefore comes from subtracting a reference defined in weight space, not from the added capacity of an offset.

\subsection{Without the Root or Tree Labels}
\label{sec:noroot}
Models collected from a hub largely belong to Model Trees: among Hugging Face models that declare a base model, ProbeX~\citep{probex} reports that trees with at least 100 models cover over $70\%$ of models. Models that do not declare a base model are not covered by this estimate. DSProbe subtracts the pretrained root of each tree, which may not be available for such models. R-DSProbe does not need it: it only requires grouping the models that share a common origin, since $W_*$ is initialized at the mean of that tree's training leaves. We initialize $W_*$ at the pretrained root, at the leaf mean, and at a random matrix (Table~\ref{tab:noroot}). Initializing at the leaf mean, which does not use the root, changes accuracy by at most $0.32$ points relative to initializing at $W_{\mathrm{pre}}$, and it matches or exceeds DSProbe on all four architectures. A random initialization is consistently worse than both, so what R-DSProbe needs is not the root itself but the grouping of models that share a common origin, from which the leaf mean is computed.

\begin{table}[t]
\caption{R-DSProbe without the pretrained root: $W_*$ is initialized at the root, the leaf mean, or a random matrix (uncapped, as in Table~\ref{tab:reference_init}). Accuracy (\%), mean $\pm$ std over five seeds.}
\label{tab:noroot}
\centering
\resizebox{\linewidth}{!}{%
\begin{tabular}{lccccc}
\toprule
\textbf{Method} & \textbf{Uses $W_{\mathrm{pre}}$} & \textbf{ResNet} & \textbf{SupViT} & \textbf{MAE} & \textbf{DINO}\\
\midrule
DSProbe ($W - W_{\mathrm{pre}}$) & $\checkmark$ & $96.82 \pm 0.14$ & $95.93 \pm 0.15$ & $83.84 \pm 0.15$ & $77.89 \pm 0.25$\\
R-DSProbe, $W_*$ init.\ at $W_{\mathrm{pre}}$ & $\checkmark$ & $97.40 \pm 0.18$ & $95.92 \pm 0.18$ & $84.71 \pm 0.25$ & $78.91 \pm 0.22$\\
R-DSProbe, $W_*$ init.\ at $W_{\mathrm{mean}}$ & $\times$ & $97.32 \pm 0.05$ & $95.95 \pm 0.07$ & $85.03 \pm 0.27$ & $79.01 \pm 0.23$\\
R-DSProbe, $W_*$ init.\ at $W_{\mathrm{rand}}$ & $\times$ & $89.33 \pm 0.74$ & $92.09 \pm 0.57$ & $82.80 \pm 0.35$ & $75.17 \pm 0.24$\\
\bottomrule
\end{tabular}%
}
\end{table}

This grouping can be recovered from the weights alone. The leaves of a tree stay close to their common origin, whereas different trees are pretrained separately, so the trees are separated in weight space. Even in the worst case, the largest distance from a leaf to its own tree's mean is $1.7$--$3.6$ times smaller than the smallest distance from a leaf to another tree's mean (Appendix~\ref{app:tree_membership}, Table~\ref{tab:tree_separation}). We pool all $2{,}988$ leaves of the three ViT-B trees (SupViT, MAE, and DINO, which share the same architecture) without their labels and run $k$-means with $k=3$ on the weights of each tree's probing layer, over five seeds. The clusters match the true trees exactly in every layer and seed, with an Adjusted Rand Index (ARI) of $1.000$, and the same holds when the clusters are built from the training leaves only and the $895$ unseen validation and test leaves are assigned to the nearest centroid, so a new model can be assigned to its tree without any label.

The trees separate cleanly on this benchmark, but membership may be wrong for other collections of models. To quantify the effect of wrong membership, we replace a fraction $p$ of each tree's training leaves with the leaves of other trees that are closest to that tree's mean, which are the easiest to confuse. We initialize $W_*$ at the mean of the resulting leaves, train as usual, and evaluate on clean test models (Appendix~\ref{app:tree_membership}, Table~\ref{tab:tree_contamination}). Accuracy drops by $2.3$--$2.6$ points at $p=5\%$ and by $3.6$--$5.3$ points at $p=10\%$. R-DSProbe therefore relies on accurate tree membership, which the weight-based clustering provides on this benchmark.

\subsection{Efficiency of R-DSProbe}
So far, our experiments fixed the number of probes at $r{=}128$ to isolate the
effect of the input representation. We now turn to efficiency. R-DSProbe adds
a single shared reference $W_*$ per tree, increasing the number of trained
parameters, but this overhead is only about $1\%$ of the metanetwork and barely
increases training compute (Appendix~\ref{app:wstar_overhead}). The more important
axis for efficiency is the number of probes $r$. Probing is, by design, an approach
that replaces a dense metanetwork, characterizing a layer from the responses to a
few probes and thereby reducing computation~\citep{probex}. Reaching the same
accuracy with fewer probes therefore means that larger weight matrices can be
handled more efficiently. From this perspective, we vary the number of probes $r$
and measure how many probes R-DSProbe needs to match MVProbe's accuracy: R-DSProbe
reproduces the accuracy that MVProbe attains at its full budget with far fewer
probes, using $1.3$--$21.1\times$ fewer FLOPs depending on the architecture
(Appendix~\ref{app:probe_efficiency}).

\begin{figure}[t]
\centering
\includegraphics[width=\textwidth]{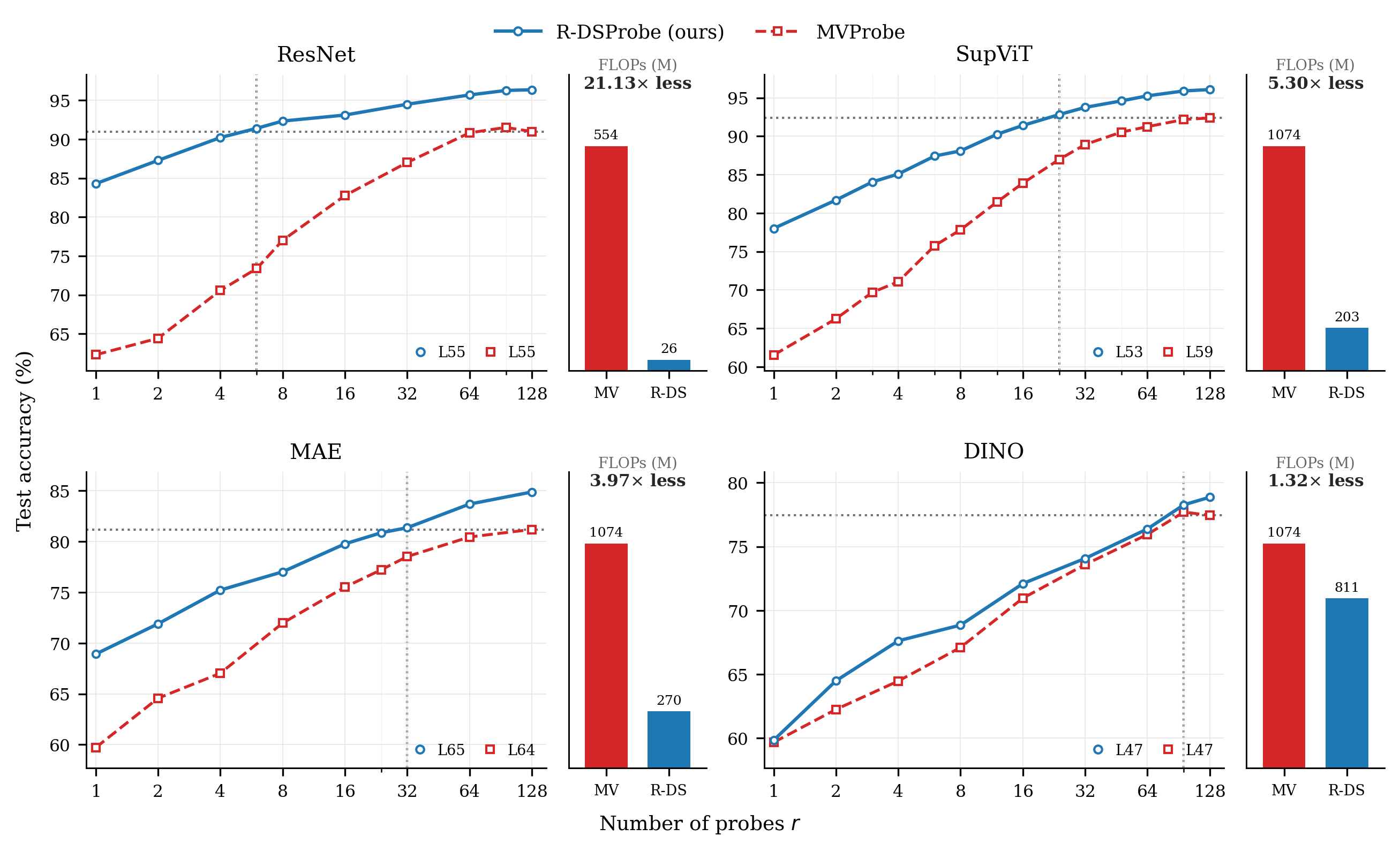}
\caption{Probe-count efficiency across four architectures. Left: test accuracy
(\%) vs.\ number of probes $r$ for R-DSProbe (blue) and MVProbe (red). Right:
matrix-multiplication FLOPs of the full probing path, including the per-batch
reference products, for MVProbe at $r{=}128$ versus R-DSProbe at the smallest $r$
matching its accuracy.}
\label{fig:probe_efficiency_full}
\end{figure}

\Figref{fig:probe_efficiency_full} shows the accuracy as a function of $r$. R-DSProbe exceeds MVProbe across the entire range on all four architectures and reaches the accuracy that MVProbe attains at its full budget ($r{=}128$) with $r{=}6$ (ResNet), $24$ (SupViT), $32$ (MAE), and $96$ (DINO), which is $21.13\times$, $5.30\times$, $3.97\times$, and $1.32\times$ fewer FLOPs. This also shortens training: at these probe counts, training takes $2.57\times$, $2.48\times$, $2.23\times$, and $1.14\times$ less wall-clock time than full-budget MVProbe under identical hardware and settings (Table~\ref{tab:wallclock}).

\subsection{Cross-Tree Probing with a Shared Encoder}
\label{sec:crosstree}

Models in practice span multiple Model Trees. ProbeX~\citep{probex} handles this
with a mixture-of-tree-experts, training a separate metanetwork per tree, so the
parameter count grows linearly with the number of trees. Most of this cost is
avoidable: the encoder $g(\Phi)$ holds $95.9\%$ of the parameters
(Appendix~\ref{app:wstar_overhead}), and since the probe and projection map every
tree to a common $d_p$-dimensional representation, a single encoder and head can be
shared across trees, routing only the lightweight per-tree components (probe $\mathbf U$
and projection $V$). Applied to raw-weight probing, however, this sharing costs $2.84$ points of mean accuracy (Table~\ref{tab:crosstree}, $85.78 \to 82.94$): a raw
weight carries each tree's pretrained origin, so inputs from different trees are not
aligned to a common frame and the shared encoder suffers negative transfer.
Subtracting a per-tree reference may remove this origin at its source and align
displacements from different trees into a common frame. Consistent with this, this
cost shrinks to $1.67$ points with the pretrained origin (DSProbe,
$88.43 \to 86.76$) and is smallest with the learned reference $W_*$ (R-DSProbe), at
$1.44$ points ($89.05 \to 87.61$). Subtracting a reference thus lowers the accuracy
cost of sharing, which makes the $3.61\times$ parameter reduction of sharing
($139.47 \to 38.65$M for R-DSProbe) cheaper. Per tree, each expert shrinks from
$34.87$M to $1.26$M ($28\times$), and the saving grows with the number of trees.

\begin{table}[t]
\caption{Cross-tree probing with a shared encoder, crossing the probe input with the encoder. Every model is trained uncapped. Values are mean
$\pm$ population standard deviation over five seeds.}
\label{tab:crosstree}
\centering
\resizebox{\linewidth}{!}{%
\begin{tabular}{lcccccc}
\toprule
\textbf{Method} & \textbf{ResNet(L55)} & \textbf{SupViT(L59)} & \textbf{MAE(L64)} & \textbf{DINO(L47)} & \textbf{Mean} & \textbf{Params}\\
\midrule
Per-tree experts (raw)  & $91.73 \pm 0.38$ & $92.53 \pm 0.30$ & $81.26 \pm 0.27$ & $77.59 \pm 0.17$ & $85.78$ & $137.44$M\\
Per-tree experts (DSProbe) & $96.50 \pm 0.33$ & $95.78 \pm 0.20$ & $83.84 \pm 0.23$ & $77.60 \pm 0.25$ & $88.43$ & $137.44$M\\
Per-tree experts (R-DSProbe) & $\mathbf{96.51} \pm 0.13$ & $\mathbf{96.20} \pm 0.14$ & $\mathbf{84.74} \pm 0.12$ & $\mathbf{78.76} \pm 0.16$ & $\mathbf{89.05}$ & $139.47$M\\
Shared encoder (raw)    & $89.55 \pm 0.11$ & $91.78 \pm 0.42$ & $74.41 \pm 1.35$ & $76.03 \pm 0.29$ & $82.94$ & $36.62$M\\
Shared encoder (DSProbe) & $96.11 \pm 0.11$ & $93.88 \pm 0.37$ & $81.75 \pm 0.10$ & $75.30 \pm 0.16$ & $86.76$ & $36.62$M\\
Shared encoder (R-DSProbe)   & $\mathbf{96.14} \pm 0.20$ & $\mathbf{94.32} \pm 0.22$ & $\mathbf{83.47} \pm 0.37$ & $\mathbf{76.52} \pm 0.27$ & $\mathbf{87.61}$ & $38.65$M\\
\bottomrule
\end{tabular}%
}
\end{table}

\section{Discussion}
\label{sec:discussion}

Our results point to a simple principle: in weight-space probing, \emph{what you subtract matters as much as what you probe}. Subtracting the shared pretrained origin already turns raw weights into a more readable signal, and learning the reference beyond it
adds a further, smaller gain---prior methods read the raw weight, which is just the displacement from a zero reference. The same lens explains second-order views: higher orders separate ever more matrices yet do not improve over the Gram, so expressivity cannot be the reason. What distinguishes the Gram is what training writes into it---to first order, its change along fine-tuning records the features that fine-tuning induced, not a more expressive encoding of the same weight. Probing views should thus be chosen for what training writes into them, and references for what they remove.

Several controls confirm that the reference is what drives the gain. Evaluated in isolation with a single-branch model, the displacement improves nearly every response, so the effect lies in the input representation, not a particular combination of branches (Appendix~\ref{app:single_branch}). Nor does it come from extra capacity: a free, more expressive response offset fails to recover it (\Secref{sec:reference} and Appendix~\ref{app:offset}). And it is stable across the probed layer (Appendix~\ref{app:robustness}), given a reference initialized near the shared weight structure with capped updates (Appendix~\ref{app:reference_init}).

A one-line change of input thus becomes a different kind of probing method. R-DSProbe leaves the probing architecture untouched, yet sets a new state of the art on every architecture we study while using a fraction of the probes and a single encoder where prior work forces one expert per tree. The choice of reference, not added capacity, is what drives these gains: much of the signal lies in how the weights have moved from a reference rather than in the weights themselves.

\section{Conclusion}
\label{sec:conclusion}

We introduced the \emph{DiSplacement Probe (DSProbe)}, which reads a layer's displacement
from a reference rather than the layer itself, and traced the success of
within-tree probing to the shared pretrained origin that this displacement
removes. Extended to second-order Gram responses, the displacement tracks, to first order, the feature changes induced by fine-tuning, offering an account of their value where expressivity alone does not suffice; treating the reference as a design choice, the
\emph{Refined-DSProbe} learns the shared component to subtract and achieves
state-of-the-art accuracy on all Model Jungle architectures; it does so with
fewer probes, robustness to layer selection, and a single encoder shared
across trees.

Several limitations remain. Our analysis is first-order in the learning rate and
assumes fine-tuning from the shared root; its behavior under long schedules,
adaptive optimizers, or multi-stage training is open, and richer references than
a single shared matrix are a natural extension. Our evaluation also covers vision
backbones of moderate size, whereas displacement probing should matter most at
larger scale: language models and their fine-tuned descendants form vast Model
Trees with far larger weight matrices, where analyzing each model with less capacity becomes increasingly valuable. We consider this the most promising
direction, and hope that treating the reference as a first-class design axis
proves useful for weight-space learning more broadly.

\section*{AI Use Statement}
Large language models (LLMs) were used in the preparation of this paper as
follows. \emph{Writing:} LLMs assisted in drafting and polishing sections of the
main text and appendix, and in \LaTeX{} formatting; all drafts were revised and
verified by the authors. \emph{Retrieval:} LLMs were used to help locate related
work, which the authors read and verified independently. \emph{Proofs:} the
theorems and their proofs were formulated by the authors; LLMs were used to check
the derivations and to typeset them, and the authors verified every step.
\emph{Code:} LLMs assisted in cleaning and documenting the experimental code
prepared for public release; the code used to produce the reported results was
written and run by the authors.

\section*{Ethics Statement}
This work studies predicting model attributes from the weights of publicly
released checkpoints, using the public Model Jungle benchmark; it involves no
human subjects or private data. Because the method infers properties of a model's
training data, it could in principle reveal information a model owner did not
document; we view this primarily as a tool for improving the transparency of
undocumented public models, and note that it recovers only coarse attributes
(e.g., class membership) within a known Model Tree.

\section*{Reproducibility Statement}
All experiments use the public Model Jungle benchmark~\citep{probex}.
Code to
reproduce our results, including training scripts and configurations for all
methods, is available in the repository linked in the abstract.
Method details are given in \Secref{sec:method}, the experimental protocol
in \Secref{sec:experiments}, and the full architecture and hyperparameters
in Appendix~\ref{app:setting}. Training details and ablations for the learned
reference $W_*$ are in Appendix~\ref{app:reference_init}, the layer enumeration
used in all layer-wise analyses in Appendix~\ref{app:layer_ordering}, and
complete proofs of the theoretical results in Appendix~\ref{app:proof_higher_order}.
All results are reported as mean $\pm$ standard deviation over five seeds.

\bibliographystyle{iclr2027_conference}
\bibliography{references}

\appendix

\section{Accumulation of the Displacement}
\label{app:task_vec_update}

Let $\theta_t=\{W_{l,t}\}_{l=1}^L$ denote the model parameters at step $t$, with $W_{l,t}\in\R^{m\times n}$ the selected layer, and let $\mathcal{L}_\tau$ denote the task loss (\Secref{subsec:wsl_bg}). Under gradient descent, the selected layer is updated by
\begin{equation}
    W_{l,t+1}
    =
    W_{l,t}
    -\eta\,\nabla_{W_{l,t}}\mathcal{L}_\tau(\theta_t).
    \label{eq:layer_gd_update}
\end{equation}
The per-step displacement $-\eta\,\nabla_{W_{l,t}}\mathcal{L}_\tau(\theta_t)$ is evaluated at the current model state $\theta_t$ and measures its mismatch from the target function $f_\tau$. Summing \eqref{eq:layer_gd_update} over the full trajectory gives
\begin{equation}
    W_l = W_{l,0} + \Delta_W,
    \qquad
    \Delta_W = -\eta\sum_{t=0}^{T-1}\nabla_{W_{l,t}}\mathcal{L}_\tau(\theta_t),
    \label{eq:endpoint_task_update}
\end{equation}
where $W_l:=W_{l,T}$. Thus $\Delta_W$ is the accumulated task-conditioned displacement, formed by gradients evaluated along the full-model trajectory $\{\theta_t\}_{t=0}^{T-1}$.

Within a Model Tree, the shared pretrained root $W_{\mathrm{pre}}$ replaces the initial weight $W_{l,0}$,
\begin{equation}
    \theta_0^{(\tau)} = \theta_{\mathrm{pre}},
    \qquad
    W_{l,0}^{(\tau)} = W_{\mathrm{pre}},
    \qquad \forall \tau.
    \label{eq:shared_initialization}
\end{equation}
Indexing each leaf $W_{l,T}^{(\tau)}$ by its source task $\tau$, the root cancels already in the first-step difference of two leaves, leaving only the difference of task-specific loss gradients, both evaluated at the same root state $\theta_0=\theta_{\mathrm{pre}}$:
\begin{equation}
    W_{l,1}^{(\tau)} - W_{l,1}^{(\tau')}
    =
    -\eta\left[
        \nabla_{W_{l,0}}\mathcal{L}_\tau(\theta_0)
        - \nabla_{W_{l,0}}\mathcal{L}_{\tau'}(\theta_0)
    \right]
    =
    -\eta\,\nabla_{W_{l,0}}\left(\mathcal{L}_\tau - \mathcal{L}_{\tau'}\right)(\theta_0).
    \label{eq:initial_task_differentiation}
\end{equation}
At the first step all leaves share the same parameter state, so each task induces a distinct gradient direction in the shared parameter space. After fine-tuning, every leaf can be written as $W_{l,T}^{(\tau)} = W_{\mathrm{pre}} + \Delta_W^{(\tau)}$, so the shared pretrained component cancels when comparing leaves, and the inter-leaf differences are entirely determined by the displacements $\Delta_W^{(\tau)}$.

\section{Gram Feature Updates}
\label{app:gram_feature_update}

We derive the two relations in \eqref{eq:gram_feature_updates}. Consider the selected layer $l$ at optimization step $t$,
\begin{equation}
    Z_t = W_{l,t} H_t,
\end{equation}
where $H_t$ and $Z_t$ denote its input and output features, respectively.
Since $\mathcal{L}_\tau$ is scalar-valued, the chain rule gives
\begin{equation}
    \nabla_{W_{l,t}}\mathcal{L}_\tau(\theta_t)
    =
    \nabla_{Z_t}\mathcal{L}_\tau(\theta_t)\,H_t^\top,
    \qquad
    \nabla_{H_t}\mathcal{L}_\tau(\theta_t)
    =
    W_{l,t}^\top
    \nabla_{Z_t}\mathcal{L}_\tau(\theta_t).
    \label{eq:gram_chain_rule}
\end{equation}
Consequently,
\begin{equation}
    W_{l,t}^\top
    \nabla_{W_{l,t}}\mathcal{L}_\tau(\theta_t)
    =
    \nabla_{H_t}\mathcal{L}_\tau(\theta_t)\,H_t^\top,
    \label{eq:gram_feature_gradient}
\end{equation}
and, using $Z_t=W_{l,t}H_t$,
\begin{equation}
    \nabla_{W_{l,t}}\mathcal{L}_\tau(\theta_t)
    W_{l,t}^\top
    =
    \nabla_{Z_t}\mathcal{L}_\tau(\theta_t)\,Z_t^\top.
    \label{eq:gram_output_gradient}
\end{equation}

We first consider the input-side Gram $W_{l,t}^\top W_{l,t}$.
Under the gradient update
\[
    W_{l,t+1}
    =
    W_{l,t}
    -
    \eta\,\nabla_{W_{l,t}}\mathcal{L}_\tau(\theta_t),
\]
its one-step change is
\begin{align}
    W_{l,t+1}^\top W_{l,t+1}
    &=
    \left(
        W_{l,t}
        -
        \eta\,\nabla_{W_{l,t}}\mathcal{L}_\tau(\theta_t)
    \right)^\top
    \left(
        W_{l,t}
        -
        \eta\,\nabla_{W_{l,t}}\mathcal{L}_\tau(\theta_t)
    \right)
    \nonumber\\
    &=
    W_{l,t}^\top W_{l,t}
    -
    \eta\left[
        W_{l,t}^\top
        \nabla_{W_{l,t}}\mathcal{L}_\tau(\theta_t)
        +
        \nabla_{W_{l,t}}\mathcal{L}_\tau(\theta_t)^\top
        W_{l,t}
    \right]
    \nonumber\\
    &\quad
    +
    \eta^2
    \nabla_{W_{l,t}}\mathcal{L}_\tau(\theta_t)^\top
    \nabla_{W_{l,t}}\mathcal{L}_\tau(\theta_t).
\end{align}
Therefore,
\begin{align}
    W_{l,t+1}^\top W_{l,t+1}
    -
    W_{l,t}^\top W_{l,t}
    &=
    -2\eta\,
    \operatorname{Sym}\!\left(
        W_{l,t}^\top
        \nabla_{W_{l,t}}\mathcal{L}_\tau(\theta_t)
    \right)
    \nonumber\\
    &\quad
    +
    \eta^2
    \nabla_{W_{l,t}}\mathcal{L}_\tau(\theta_t)^\top
    \nabla_{W_{l,t}}\mathcal{L}_\tau(\theta_t)
    \nonumber\\
    &=
    -2\eta\,
    \operatorname{Sym}\!\left(
        \nabla_{H_t}\mathcal{L}_\tau(\theta_t)\,H_t^\top
    \right)
    +
    \mathcal{O}(\eta^2),
    \label{eq:weight_gram_input_update}
\end{align}
where $\operatorname{Sym}(A):=(A+A^\top)/2$.

The output-side Gram $W_{l,t}W_{l,t}^\top$ follows analogously:
\begin{align}
    W_{l,t+1}W_{l,t+1}^\top
    &=
    \left(
        W_{l,t}
        -
        \eta\,\nabla_{W_{l,t}}\mathcal{L}_\tau(\theta_t)
    \right)
    \left(
        W_{l,t}
        -
        \eta\,\nabla_{W_{l,t}}\mathcal{L}_\tau(\theta_t)
    \right)^\top
    \nonumber\\
    &=
    W_{l,t}W_{l,t}^\top
    -
    \eta\left[
        W_{l,t}
        \nabla_{W_{l,t}}\mathcal{L}_\tau(\theta_t)^\top
        +
        \nabla_{W_{l,t}}\mathcal{L}_\tau(\theta_t)
        W_{l,t}^\top
    \right]
    \nonumber\\
    &\quad
    +
    \eta^2
    \nabla_{W_{l,t}}\mathcal{L}_\tau(\theta_t)
    \nabla_{W_{l,t}}\mathcal{L}_\tau(\theta_t)^\top.
\end{align}
Hence,
\begin{align}
    W_{l,t+1}W_{l,t+1}^\top
    -
    W_{l,t}W_{l,t}^\top
    &=
    -2\eta\,
    \operatorname{Sym}\!\left(
        \nabla_{W_{l,t}}\mathcal{L}_\tau(\theta_t)
        W_{l,t}^\top
    \right)
    \nonumber\\
    &\quad
    +
    \eta^2
    \nabla_{W_{l,t}}\mathcal{L}_\tau(\theta_t)
    \nabla_{W_{l,t}}\mathcal{L}_\tau(\theta_t)^\top
    \nonumber\\
    &=
    -2\eta\,
    \operatorname{Sym}\!\left(
        \nabla_{Z_t}\mathcal{L}_\tau(\theta_t)\,Z_t^\top
    \right)
    +
    \mathcal{O}(\eta^2).
    \label{eq:weight_gram_output_update}
\end{align}

Thus, to first order in the learning rate, changes in the two weight-Gram orientations correspond to covariance-like shifts in the layer input and output feature spaces, respectively. In particular, the first relation is expressed through the task-conditioned input-feature gradient $\nabla_{H_t}\mathcal{L}_\tau$, whereas the second is expressed through the corresponding output-feature gradient $\nabla_{Z_t}\mathcal{L}_\tau$. This yields the two relations stated in \eqref{eq:gram_feature_updates}.

The single-step Gram shifts accumulate along the fine-tuning trajectory. Summing the first-order updates gives
\begin{equation}
    (W_l^{(\tau)})^\top W_l^{(\tau)}
    = W_{l,0}^\top W_{l,0} + \Delta_H^{(\tau)},\qquad
    W_l^{(\tau)}(W_l^{(\tau)})^\top
    = W_{l,0}W_{l,0}^\top + \Delta_Z^{(\tau)},
\label{eq:gram_endpoints}
\end{equation}
where $\Delta_H^{(\tau)}$ and $\Delta_Z^{(\tau)}$ denote the accumulated task-conditioned feature-space shifts along the optimization trajectory. With $W_{l,0}=W_{\mathrm{pre}}$, these are the Gram displacements $W_l^\top W_l - W_{\mathrm{pre}}^\top W_{\mathrm{pre}}$ and $W_lW_l^\top - W_{\mathrm{pre}}W_{\mathrm{pre}}^\top$ probed by DSProbe.

\section{Implementation Details}
\label{app:setting}
\subsection{Environment}
All experiments use four NVIDIA RTX 6000 Ada GPUs with a batch size of 128. Independent model--layer--seed runs are distributed across GPUs. We evaluate all experiments over five random seeds and report the mean performance.
\subsection{Architecture Details}
To ensure a controlled comparison of the input representations, we use the same probe network architecture as MVProbe~\citep{mvprobe}.
Let $W_l\in \R^{m_l\times n_l}$ denote a selected layer, with $n_l$ and $m_l$ its input and output dimensions.

The probe network consists of four parallel branches $\{\Phi_i\}_{i=1}^4$, each operating on a different response derived from $W_l$.
Each branch contains its own set of probe vectors $\mathbf U_i$ and an encoder that maps the corresponding response to a fixed-dimensional representation. The four branch representations are then concatenated and passed through a shared encoder and prediction head.
\paragraph{Branch configuration} Each branch has its own learnable probe matrix and a bias-free projection layer. Each branch response is first standardized per sample, $\tilde{\Phi}_i = \frac{\Phi_i - \mu(\Phi_i)}{\sigma(\Phi_i) + \epsilon}$, where $\mu$ and $\sigma$ are taken over all entries of the response. The projection then maps each of the $r$ probe responses to a $d_p$-dimensional vector, followed by a ReLU, and the $r$ projected responses are flattened into an $rd_p$-dimensional branch representation. The branches below are written for the raw weight $W_l$, as in MVProbe. DSProbe and R-DSProbe keep every shape and component unchanged and only replace each branch input by its displacement counterpart, $\Delta$ for DSProbe and $\widetilde{\Delta}$ for R-DSProbe: $W_l \to \Delta_W$ for $\Phi_1,\Phi_2$, $W_l^\top W_l \to \Delta_H$ for $\Phi_3$, and $W_l W_l^\top \to \Delta_Z$ for $\Phi_4$. Note that the second-order inputs are differences of Grams, not the Gram of the first-order displacement.

\begin{minipage}[t]{0.48\linewidth}
\begin{itemize}[leftmargin=*]
\item \textbf{Branch $\Phi_1$}
\begin{itemize}[leftmargin=*]
    \item Input: $W_l \in \mathbb{R}^{m_l \times n_l}$
    \item Probe: $\mathbf{U}_1 \in \mathbb{R}^{n_l \times r}$
    \item Response: $W_l\mathbf{U}_1 \in \mathbb{R}^{m_l \times r}$
    \item Projection and Flattening: $\mathbb{R}^{m_l \times r} \rightarrow \mathbb{R}^{r d_p}$
\end{itemize}

\item \textbf{Branch $\Phi_2$}
\begin{itemize}[leftmargin=*]
    \item Input: $W_l^\top \in \mathbb{R}^{n_l \times m_l}$
    \item Probe: $\mathbf{U}_2 \in \mathbb{R}^{m_l \times r}$
    \item Response: $W_l^\top\mathbf{U}_2 \in \mathbb{R}^{n_l \times r}$
    \item Projection and Flattening: $\mathbb{R}^{n_l \times r} \rightarrow \mathbb{R}^{r d_p}$
\end{itemize}
\end{itemize}
\end{minipage}
\hfill
\begin{minipage}[t]{0.48\linewidth}
\begin{itemize}[leftmargin=*]
\item \textbf{Branch $\Phi_3$}
\begin{itemize}[leftmargin=*]
    \item Input: $W_l^\top W_l \in \mathbb{R}^{n_l \times n_l}$
    \item Probe: $\mathbf{U}_3 \in \mathbb{R}^{n_l \times r}$
    \item Response: $W_l^\top W_l\mathbf{U}_3 \in \mathbb{R}^{n_l \times r}$
    \item Projection and Flattening: $\mathbb{R}^{n_l \times r} \rightarrow \mathbb{R}^{r d_p}$
\end{itemize}

\item \textbf{Branch $\Phi_4$}
\begin{itemize}[leftmargin=*]
    \item Input: $W_l W_l^\top \in \mathbb{R}^{m_l \times m_l}$
    \item Probe: $\mathbf{U}_4 \in \mathbb{R}^{m_l \times r}$
    \item Response: $W_l W_l^\top\mathbf{U}_4 \in \mathbb{R}^{m_l \times r}$
    \item Projection and Flattening: $\mathbb{R}^{m_l \times r} \rightarrow \mathbb{R}^{r d_p}$
\end{itemize}
\end{itemize}
\end{minipage}

\paragraph{Linear Classifier}
The four branch representations are concatenated into a \(4rd_p\)-dimensional feature vector and mapped to a 512-dimensional representation by a shared linear encoder. A linear classification head then maps this representation to class logits.

\subsection{Hyperparameters}
The hyperparameters used in our experiments are in Table~\ref{tab:hyperparams}. We use the ProbeX~\citep{probex} architecture with $d=128$ to evaluate individual displacement responses. For the generative Model Jungle setting, we map frozen CLIP text embeddings to the same 512-dimensional space as the probe representation through a learnable linear projection. Cosine similarities between the probe representation and the projected class embeddings serve as classification logits.

\begin{table*}[t]
\centering
\caption{Hyperparameters used in our experiments.}
\label{tab:hyperparams}
\resizebox{\linewidth}{!}{%
\begin{tabular}{lc@{\qquad}lc@{\qquad}lc}
\toprule
\multicolumn{2}{c}{\textbf{DSProbe / MVProbe}}
&
\multicolumn{2}{c}{\textbf{ProbeX / ProbeX($\times4$)}}
&
\multicolumn{2}{c}{\textbf{$W_*$ Optimization}} \\
\cmidrule(lr){1-2}
\cmidrule(lr){3-4}
\cmidrule(lr){5-6}

\textbf{Hyperparameter} & \textbf{Value}
& \textbf{Hyperparameter} & \textbf{Value}
& \textbf{Hyperparameter} & \textbf{Value} \\
\midrule

Epochs & 500
& Epochs & 200 / 500
& Learning rate & $3\times10^{-4}$ \\

Learning rate & $3\times10^{-4}$
& Learning rate & $2\times10^{-3}$ / $3\times10^{-4}$
& Weight decay & $10^{-5}$ \\

Weight decay & $10^{-5}$
& Weight decay & $0$ / $10^{-5}$
& Optimizer & Adam \\

Batch size & 128
& Batch size & 128
& Update cap & $10^{-3}$ \\

Probes per branch ($r$) & 128
& Probes per branch ($r$) & 128 / 512
& & \\

Projection ($d$) / representation & $128 / 512$
& Projection ($d$) / representation & $128 / 512$
& & \\

Branches & 4
& Branches & 1
& & \\

Response standardization & True
& Response standardization & None
& & \\

Optimizer & Adam
& Optimizer & Adam
& & \\

Loss & BCE
& Loss & BCE
& & \\

\bottomrule
\end{tabular}%
}
\end{table*}

\section{Experiment Details}
\label{app:experiment_details}
\subsection{Datasets}
Model Jungle~\citep{probex} comprises collections of randomly fine-tuned models derived from common foundation models, emulating the Model Tree setting. We consider two evaluation settings:
\begin{itemize}
    \item \textbf{Discriminative.} Each model is fine-tuned on a random subset of 50 CIFAR-100~\citep{cifar100} classes, and the task is to predict source-class membership. We evaluate four architectures: ResNet-101~\citep{resnet}, SupViT~\citep{supvit}, MAE~\citep{mae}, DINO~\citep{dino}.
    \item \textbf{Generative.} Each Stable Diffusion~\citep{ldm} LoRA~\citep{LoRA} model is fine-tuned on images sampled from a single ImageNet-1k~\citep{imagenet} class, using base checkpoints from SD-v1.2 through SD-v1.5. We reconstruct the LoRA update as $W_l = BA$ and predict the training class by mapping the model weights to the corresponding CLIP~\citep{clip} text embedding. Because $BA$ is itself the displacement from the pretrained base weight, the pretrained reference is zero in these coordinates and DSProbe coincides with MVProbe; since only $BA$ is probed, the differing base checkpoints also do not enter the input. This setting therefore evaluates only the refinement stage, with $W_*$ initialized at the mean of the training LoRA updates.
    We split the classes into 170 seen and 30 unseen classes for SD-200, and 850 seen and 150 unseen classes for SD-1k, evaluating ID and held-out-class accuracy within their respective candidate sets.
\end{itemize}

\subsection{Baselines} We compare DSProbe with three representative weight-space learning methods. Our StatNN baseline adapts the summary-statistic representation of \citet{unterthiner} to source-class membership prediction. For each parameter tensor, we compute seven statistics: the mean, population standard deviation, and the 0th, 25th, 50th, 75th, and 100th percentiles. These statistics are concatenated in parameter-name order across all named parameter tensors of the full model, including weights and biases but excluding buffers such as BatchNorm~\citep{batchnorm} running statistics. The resulting features are used to train an XGBoost~\citep{xgboost} classifier with 100 binary membership outputs. ProbeX probes individual layer weight matrices directly. MVProbe~\citep{mvprobe} augments single-layer probing with transposed and Gram-based second-order responses.

We exclude graph-based and other equivariant weight-space methods~\citep{zhang2023neural, dwsnets, neuralgraphs, gmn} from the main experiments because they are computationally prohibitive for the large weight matrices considered here and have been reported to diverge in the single-layer probing setting~\citep{probex}. We also exclude ProbeGen~\citep{probegen}, which passes probes through the entire model and therefore does not apply to probing a single layer~\citep{probex}.

\subsection{Implementation Details}
To minimize architectural confounding, we match the branch capacity: every probing method uses 512 probe vectors in total (ProbeX places all
512 on its single response, the four-branch methods use $r=128$ per response), a
bias-free per-probe projection to $d_p=128$ features, and a single linear encoder
onto a 512-dimensional representation. Unless stated otherwise, all probing methods are trained for 500 epochs with Adam (learning rate \(3\times10^{-4}\), weight decay \(10^{-5}\)) and batch size 128, as summarized in Table~\ref{tab:hyperparams}.
StatNN instead uses XGBoost 3.2.0 with CPU histogram-based training and a binary logistic objective for each membership output. We use at most 500 boosting rounds, a learning rate of 0.05, maximum tree depth of 3, a subsampling ratio of 0.9, full feature sampling, and L1 and L2 regularization coefficients of 0 and 1, respectively. Early stopping uses validation binary log loss with a patience of 25 rounds. The selected boosting iteration is then evaluated on the test set. Accuracy is the percentage of correct binary membership predictions across all test models and 100 classes, with probabilities of at least 0.5 classified as positive. Results are averaged over five seeds.

\begin{table}[t]
\caption{
Full version of the discriminative results in Table~\ref{tab:main}.
Source-class membership accuracy (\%) on Model Jungle discriminative models.
ProbeX ($\times 4$) matches the parameter budget of the four-branch methods.
Values are mean $\pm$ population standard deviation over five seeds. Best per column in bold.
}
\label{tab:main_discriminative}
\centering
\small
\setlength{\tabcolsep}{4pt}
\vspace{3pt}
\begin{tabular}{lcccc}
\toprule
\textbf{Method}
& \textbf{ResNet}
& \textbf{SupViT}
& \textbf{MAE}
& \textbf{DINO} \\
\midrule
StatNN
& $65.834 \pm 0.316$
& $52.430 \pm 0.444$
& $50.543 \pm 0.210$
& $50.530 \pm 0.246$ \\
ProbeX ($\times 4$)
& $89.891 \pm 0.343$
& $89.167 \pm 0.417$
& $77.023 \pm 0.340$
& $69.873 \pm 0.094$ \\
MVProbe
& $91.089 \pm 0.329$
& $92.413 \pm 0.398$
& $81.238 \pm 0.195$
& $77.879 \pm 0.144$ \\
\textbf{DSProbe (ours)}
& $96.822 \pm 0.143$
& $95.930 \pm 0.147$
& $83.838 \pm 0.154$
& $77.887 \pm 0.248$ \\
\textbf{R-DSProbe (ours)}
& $\mathbf{97.330 \pm 0.051}$
& $\mathbf{96.511 \pm 0.187}$
& $\mathbf{85.243 \pm 0.265}$
& $\mathbf{79.113 \pm 0.132}$ \\
\bottomrule
\end{tabular}
\end{table}

\begin{table}[t]
\caption{
Full version of the generative results in Table~\ref{tab:main}.
Classification accuracy (\%) on Stable Diffusion LoRA models.
ProbeX ($\times 4$) matches the parameter budget of the four-branch methods.
Values are mean $\pm$ population standard deviation over five seeds. Best per column in bold.
}
\label{tab:main_generative}
\centering
\small
\setlength{\tabcolsep}{4pt}
\vspace{3pt}
\begin{tabular}{lcccc}
\toprule
& \multicolumn{2}{c}{\textbf{SD-200}}
& \multicolumn{2}{c}{\textbf{SD-1k}} \\
\cmidrule(lr){2-3}\cmidrule(lr){4-5}
\textbf{Method}
& \textbf{ID}
& \textbf{Held-out}
& \textbf{ID}
& \textbf{Held-out} \\
\midrule
ProbeX ($\times 4$)
& $97.916 \pm 0.654$
& $93.253 \pm 1.569$
& $32.224 \pm 3.437$
& $52.774 \pm 3.884$ \\
MVProbe
& $99.800 \pm 0.127$
& $95.250 \pm 1.425$
& $97.996 \pm 0.283$
& $97.086 \pm 0.430$ \\
\textbf{R-DSProbe (ours)}
& $\mathbf{99.840 \pm 0.150}$
& $\mathbf{95.529 \pm 0.753}$
& $\mathbf{99.038 \pm 0.234}$
& $\mathbf{98.483 \pm 0.240}$ \\
\bottomrule
\end{tabular}
\end{table}

For each probing method and model family we select the probing layer by validation
accuracy from the layer-wise sweep of \Figref{fig:layerwise}, in which the
ProbeX baseline follows its original single-branch protocol (200 epochs,
$2\times10^{-3}$, no weight decay); the parameter-matched ProbeX~($\times4$)
reported in Table~\ref{tab:main} is evaluated at the layers this
sweep selects. The selected layers are 98, 59, 64, and 59 for ProbeX, 55, 59,
64, and 47 for MVProbe, and 98, 53, 65, and 47 for DSProbe and R-DSProbe, for
ResNet, SupViT, MAE, and DINO respectively. The ablations that change only the
reference (Tables~\ref{tab:reference},~\ref{tab:offset},~\ref{tab:reference_init}
and~\ref{tab:reference_cap}) inherit the DSProbe layers, while the raw-weight
row of each is the MVProbe result at MVProbe's own layers, so that no method is
evaluated at a layer selected for a different one.

Three comparisons instead hold every condition at one layer set, because the
quantity being compared demands it. The branch-composition ablations
(Tables~\ref{tab:higher_order_branch_ablation}
and~\ref{tab:single_branch_response}) vary the architecture rather than the
reference and use 55, 59, 64, and 47 throughout; the cross-tree experiment
(Table~\ref{tab:crosstree}) uses the same set so that per-tree and shared models
are counted at identical layers. The probe-count comparison
(\Figref{fig:probe_efficiency_full}) requires both methods to probe matrices
of the same shape; this already holds for SupViT, MAE, and DINO, and for ResNet
both methods are evaluated at layer 55, as listed in Table~\ref{tab:eff_layers}.

For SD-LoRA we probe L46 and train with cross-entropy over the candidate classes
of each split (170 in-distribution and 30 held-out for SD-200, 850 and 150 for
SD-1k) rather than the full label set. Class representations are frozen CLIP
ViT-L/14 text embeddings of the prompt \emph{a photo of a \{class\}}, mapped into
the 512-dimensional representation space by a learned projection; cosine
similarities serve as logits. In-distribution and held-out-class checkpoints are
selected independently using their corresponding validation candidate sets. The held-out
classes of the validation and test sets largely overlap (for SD-200, all 30 held-out classes
appear in both; for SD-1k, 134 of the 150 held-out candidate classes appear in validation and
149 in testing, with 133 shared), following the protocol of MVProbe. Selecting checkpoints on
the held-out validation set therefore uses information about the held-out classes, and the
evaluation is not a strict class-agnostic zero-shot setting.
The full architecture and the procedure for training $W_*$ are described in
Appendix~\ref{app:setting} and Appendix~\ref{app:reference_init}.

\section{Higher-Order Branch Ablation}
\label{app:higher_order}
MVProbe motivates its second-order branches by the additional expressivity.
\begin{theorem}[{Expressiveness of Second-Order Probes; adapted from
MVProbe~\citep[Theorem~4.1]{mvprobe}}]
Let $\mathbf U\in\mathbb R^{n\times r}$ be a probe matrix with
$\operatorname{rank}(\mathbf U)=r<n$.
Define the first- and second-order probe responses as
\begin{equation}
\Phi_1(X;\mathbf U)
=
X\mathbf U,
\qquad
\Phi_2(X;\mathbf U)
=
(X^\top X)\mathbf U.
\end{equation}
Then there exist distinct matrices
$X_1,X_2\in\mathbb R^{m\times n}$ such that
\begin{equation}
\Phi_1(X_1;\mathbf U)
=
\Phi_1(X_2;\mathbf U),
\qquad\text{but}\qquad
\Phi_2(X_1;\mathbf U)
\neq
\Phi_2(X_2;\mathbf U).
\end{equation}
\end{theorem}

\begin{theorem}[Expressiveness of Higher-Order Probes]
\label{thm:higher_order_app}
Let $\mathbf U\in\mathbb R^{n\times r}$ be a probe matrix with
$\operatorname{rank}(\mathbf U)=r\geq1$.
For each integer $q\geq0$, define
\begin{equation}
\Phi_{2q}(X;\mathbf U)
=(X^\top X)^q\mathbf U,
\qquad
\Phi_{2q+1}(X;\mathbf U)
=X(X^\top X)^q\mathbf U.
\end{equation}
For any integer $k\geq2$, there exist distinct matrices
$X_1,X_2\in\mathbb R^{m\times n}$ such that
\begin{equation}
\Phi_j(X_1;\mathbf U)=\Phi_j(X_2;\mathbf U),
\qquad 1\leq j<k,
\end{equation}
but
\begin{equation}
\Phi_k(X_1;\mathbf U)\neq\Phi_k(X_2;\mathbf U),
\end{equation}
if and only if
\begin{equation}
n-r\geq\left\lfloor\frac{k}{2}\right\rfloor,
\qquad
m\geq\left\lceil\frac{k}{2}\right\rceil.
\end{equation}
\end{theorem}

Theorem~\ref{thm:higher_order_app} gives an exact characterization of when higher-order probe responses can expose weight-space variations that are not observable from lower-order responses alone, formalizing the expressivity argument of Section~\ref{sec:questions}. We test whether this additional expressivity translates into improved weight-space prediction. We augment MVProbe with third-order responses (\emph{Degree-3}) and with both third- and fourth-order responses (\emph{Degree-3+4}). To separate the effect of higher-order structure from the increased model capacity induced by additional branches, we additionally construct capacity-matched controls with the same number of branches but without introducing higher-order responses.

In Table~\ref{tab:higher_order_branch_ablation}, introducing higher-order branches does not consistently lead to higher weight-space prediction performance. While the Degree-3 variant improves ResNet from $91.089\%$ to $93.657\%$, the gains are not reproduced across the other model families. On SupViT, MAE, and DINO, the higher-order variants are comparable to or worse than their capacity-matched controls, and adding both third- and fourth-order branches does not provide further improvement. These results suggest that the additional expressivity of higher-order responses does not necessarily translate into more predictive representations for the target model attribute.

\begin{table}[t]
\caption{Source-class membership prediction accuracy (\%) of MVProbe with additional
third- and fourth-order branches and capacity-matched controls.
All branches are jointly trained with response standardization.
Values are mean $\pm$ population standard deviation over five seeds. Best per column in bold.
}
\label{tab:higher_order_branch_ablation}
\begin{tabular}{lcccc}
\toprule
\textbf{Method}
& \textbf{ResNet}
& \textbf{SupViT}
& \textbf{MAE}
& \textbf{DINO} \\
\midrule

MVProbe
& $91.089 \pm 0.329$
& $92.413 \pm 0.398$
& $81.238 \pm 0.195$
& $77.879 \pm 0.144$ \\

\addlinespace[3pt]

Degree-3
& $\mathbf{93.657 \pm 0.245}$
& $92.777 \pm 0.240$
& $81.117 \pm 0.296$
& $77.183 \pm 0.539$ \\

Capacity control
& $92.355 \pm 0.558$
& $92.753 \pm 0.132$
& $81.488 \pm 0.190$
& $78.628 \pm 0.205$ \\

\addlinespace[3pt]

Degree-3+4
& $93.457 \pm 0.515$
& $92.829 \pm 0.165$
& $81.362 \pm 0.468$
& $77.043 \pm 0.604$ \\

Capacity control
& $92.224 \pm 0.410$
& $\mathbf{93.240 \pm 0.278}$
& $\mathbf{82.019 \pm 0.348}$
& $\mathbf{78.722 \pm 0.096}$ \\

\bottomrule
\end{tabular}
\end{table}

\section{Overhead of the Learned Reference \texorpdfstring{$W_*$}{W*}}
\label{app:wstar_overhead}

As introduced in \Secref{subsec:refined}, R-DSProbe learns a shared
reference $W_*$: one $m\times n$ matrix per tree, adding exactly $mn$ parameters.
The reference responses are computed once per batch and shared across its examples. Their evaluation adds $12mnr$ forward matrix-multiplication FLOPs per batch, or $12mnr/B$ per example for batch size $B$. Thus, the reference parameter count is independent of $r$, while its computational cost scales linearly with $r$ and is amortized across the batch.

Table~\ref{tab:wstar_overhead} accounts for the parameters and training compute
that $W_*$ adds. We report FLOPs as twice the number of multiply--accumulates; the
values \citet{mvprobe} report as FLOPs are multiply--accumulate counts, so our
numbers are twice theirs under the same computation. Relative to the full
metanetwork---$34.4$M parameters for the ViT-based trees and $34.3$M for
ResNet---$W_*$ adds only $0.59$M and $0.26$M, about $1\%$ overall
($0.77$--$1.71\%$), and increases the training-FLOP estimate, computed as three times the forward-pass FLOPs, by less than $0.7\%$ at batch size $B=128$. The computational overhead is small because the reference responses are shared across the batch.

Table~\ref{tab:wstar_breakdown} breaks the metanetwork down by component. The
encoder dominates, at $95.9\%$ of the parameters, because it scales linearly with
the number of probes $r$ ($4 \times d_p \times r \times d_{\mathrm{rep}}$); the
probes and projections are a further $2.2\%$. This distinction becomes relevant under the probe-count
sweep of Appendix~\ref{app:probe_efficiency}: reducing $r$ shrinks the encoder,
and hence the whole metanetwork, while $W_*$ stays fixed.

\begin{table}[t]
\caption{Parameter and training-compute overhead of the learned reference $W_*$ in
R-DSProbe. $W_*$ adds one $m\times n$ matrix per tree; percentages are relative to
the full metanetwork ($r{=}128$). Training FLOPs are estimated as three times the
forward pass (one forward, two backward).}
\label{tab:wstar_overhead}
\centering
\resizebox{\columnwidth}{!}{%
\begin{tabular}{lcccc}
\toprule
\textbf{Model} & \textbf{Metanetwork} & \textbf{$+\,W_*$} & \textbf{Total (incr.)} & \textbf{Train FLOPs (incr.)}\\
\midrule
SupViT / MAE / DINO & $34.39$M & $+0.59$M & $34.98$M ($+1.71\%$) & $3222 \to 3243$M ($+0.66\%$)\\
ResNet              & $34.26$M & $+0.26$M & $34.52$M ($+0.77\%$) & $1661 \to 1671$M ($+0.57\%$)\\
\bottomrule
\end{tabular}%
}
\end{table}

\begin{table}[t]
\caption{Metanetwork parameter breakdown (SupViT, $768\times768$, $r{=}128$). The
encoder scales linearly with the probe count $r$ and dominates the parameter count,
whereas the reference $W_*$ is an $r$-independent constant.}
\label{tab:wstar_breakdown}
\centering
\begin{tabular}{lccc}
\toprule
\textbf{Component} & \textbf{Shape} & \textbf{Params} & \textbf{Share}\\
\midrule
Probes $\mathbf{U}$ (4 branches) & $4\times(768,128)$ & $0.39$M   & $1.1\%$\\
Projections (4)                  & $4\times(128,768)$ & $0.39$M   & $1.1\%$\\
Encoder                          & $(512,65536)$      & $33.55$M  & $95.9\%$\\
Classifier head                  & $(100,512)$        & $0.05$M   & $0.1\%$\\
\midrule
Reference $W_*$                  & $(768,768)$        & $0.59$M   & $1.7\%$\\
\bottomrule
\end{tabular}
\end{table}

\section{Probe-Count Efficiency}
\label{app:probe_efficiency}

R-DSProbe reuses the multi-branch probing architecture of MVProbe and changes only
the input representation to the displacement. The computational cost of both
methods is therefore dominated by the same four probe-response branches, each of
which costs $O(mnr)$~\citep{mvprobe}, where $r$ is the number of probes and $(m,n)$
the weight-matrix size. This cost depends only on $r$ and $(m,n)$, not on the
identity---depth or type---of the probed layer.
For this comparison, we count matrix-multiplication FLOPs for the full probing
path---the weight responses, branch projections, encoder, prediction head, and the
per-batch reference products. The two methods share every term of this path except
the reference products, so at the same $r$ they differ only by the batch-amortized
$12mnr/B$ term charged to R-DSProbe ($0.6$--$0.7\%$ at $B{=}128$, $r{=}128$;
Appendix~\ref{app:wstar_overhead}) whenever the probed layers have the same
$(m,n)$, even if the layers themselves differ. We exploit this to construct a sweep
that is iso-cost up to this term: each method is evaluated at its own
validation-selected layer, and we compare only when the two selected layers share
the same size, so that the sweep varies the probing capacity $r$ alone.

Table~\ref{tab:eff_layers} lists the layers used in the comparison and their weight
sizes. For SupViT, MAE, and DINO, the validation-selected layers of the two methods
share the same $768\times768$ size and thus satisfy the iso-cost condition. For
ResNet, the two methods' optimal layers differ in size (R-DSProbe: L98, MVProbe:
L55), so their FLOPs would not match at a given $r$; to preserve iso-cost we
evaluate both methods at MVProbe's optimal layer (L55, $1024\times256$), an
unfavorable setting for R-DSProbe since it is not evaluated at its own best layer. In
every sweep we vary only the number of probes $r$ and keep all other training
settings identical to the main experiments, computing the metanetwork's FLOPs at
each $r$ from the weight size of the probed layer.

The same weight-matrix sizes are used for the parameter and
training-compute accounting in Table~\ref{tab:wstar_overhead}, with $r=128$. The reported
parameter overhead of $0.77$--$1.71\%$ therefore applies to these
fixed-shape configurations. \Figref{fig:probe_efficiency_full} shows the accuracy--$r$ curve for each
architecture. R-DSProbe exceeds MVProbe across the entire range of $r$ on all four
architectures, and reaches the accuracy that MVProbe attains at its full budget
($r{=}128$) with far fewer probes: $r{=}6$ (ResNet), $24$ (SupViT), $32$ (MAE), and
$96$ (DINO). Because the two methods share the same weight size within each panel,
this reduction in probes translates directly into a reduction in compute, shown by
the bars beside each panel: $21.13\times$, $5.30\times$, $3.97\times$, and
$1.32\times$ fewer FLOPs for ResNet, SupViT, MAE, and DINO, respectively.

Because the encoder scales linearly with $r$ while $W_*$ does not
(Appendix~\ref{app:wstar_overhead}), these probe reductions also shrink the
metanetwork itself. Reducing SupViT to $r{=}24$ cuts the metanetwork from $34.98$M
to $7.40$M parameters ($4.7\times$), and ResNet to $r{=}6$ from $34.52$M to
$2.23$M ($15.5\times$); in each case, the reference contributes a fixed number of
parameters independent of $r$: $0.59$M for SupViT and $0.26$M for ResNet. The
learned reference is therefore a fixed cost that displacement probing more than
offsets by reaching the same accuracy at a much smaller probe budget. Although the
relative parameter share of $W_*$ grows as $r$ decreases, its parameter count
remains unchanged ($0.59$M / $0.26$M). The FLOP reductions in
\Figref{fig:probe_efficiency_full} include the per-batch reference computation, and
the corresponding parameter overhead is detailed in Table~\ref{tab:wstar_overhead}.
These counts are analytic and require no model bank; the released code reproduces
them directly from the layer shapes and probe counts.

We extend the FLOPs analysis to the actual training time. We compare full-budget MVProbe ($r{=}128$) with R-DSProbe at the reduced probe counts needed to match its accuracy, under identical hardware and training settings (Table~\ref{tab:wallclock}). The FLOPs savings translate into training speedups of $1.14$--$2.57\times$, and training takes less than one minute on ResNet, SupViT, and MAE.

\begin{table}[t]
\caption{Training efficiency of MVProbe and R-DSProbe at accuracy-matched probe budgets. Training time in minutes (mean $\pm$ standard deviation). Probe reductions are relative to full-budget MVProbe.}
\label{tab:wallclock}
\centering
\resizebox{\linewidth}{!}{%
\begin{tabular}{lcccc}
\toprule
\textbf{Model} & \textbf{Probes (MVProbe $\to$ R-DSProbe)} & \textbf{Probe reduction} & \textbf{Training time (min), MVProbe $\to$ R-DSProbe} & \textbf{Speedup}\\
\midrule
ResNet & $128 \to 6$ & $95.31\%$ & $0.908 \pm 0.026 \to 0.354 \pm 0.024$ & $2.569\times$\\
SupViT & $128 \to 24$ & $81.25\%$ & $1.393 \pm 0.005 \to 0.562 \pm 0.006$ & $2.477\times$\\
MAE & $128 \to 32$ & $75.00\%$ & $1.381 \pm 0.010 \to 0.619 \pm 0.009$ & $2.231\times$\\
DINO & $128 \to 96$ & $25.00\%$ & $1.310 \pm 0.002 \to 1.145 \pm 0.010$ & $1.144\times$\\
\bottomrule
\end{tabular}%
}
\end{table}

\begin{table}[t]
\caption{Layers used in the probe-count efficiency comparison and their weight
sizes. For SupViT, MAE, and DINO each method is evaluated at its own
validation-selected layer; these layers share the same size $(m,n)$; for ResNet the two
optimal layers differ in size, so both methods are evaluated at MVProbe's optimal
layer (an unfavorable setting for R-DSProbe).}
\label{tab:eff_layers}
\centering
\begin{tabular}{lcccc}
\toprule
\textbf{Model} & \textbf{MVProbe}  & \textbf{DSProbe} & \textbf{R-DSProbe} & \textbf{Size $(m,n)$}\\
\midrule
ResNet & L55 & L55 & L55 & $1024\times256$\\
SupViT & L59 & L53 & L53 & $768\times768$\\
MAE    & L64 & L65 & L65 & $768\times768$\\
DINO   & L47 & L47 & L47 & $768\times768$\\
\bottomrule
\end{tabular}
\end{table}

\section{Comparing Displacements and Their Counterparts}
\label{app:single_branch}

To evaluate how informative each response is for weight-space prediction, we evaluate each response independently using the same single-branch architecture. For each response, we compare the raw weight response and its displacements:
$W_l$, $\Delta_W$, $\widetilde{\Delta}_W$ for the first-order response;
$W_lW_l^\top$, $\Delta_Z$, $\widetilde{\Delta}_Z$ and
$W_l^\top W_l$, $\Delta_H$, $\widetilde{\Delta}_H$ for the second-order responses.
All nine variants share the probing capacity and the optimization settings: the
probe count, the projection and representation widths, the number of epochs, the
optimizer, the learning rate, the weight decay and the batch size. The
response-standardization and probe-initialization protocol instead follows the
response type, as in the main experiments: the first-order conditions use the
original unstandardized ProbeX protocol, while the Gram conditions use response
standardization and Xavier probe initialization. The three variants within a
response type are therefore identical except for the input representation, which
is the comparison this table makes; accuracies are compared within a response
type rather than across them.

Table~\ref{tab:single_branch_response} shows that displacement probing generally improves weight-space prediction performance. The pretrained displacement improves 10 of the 12 model--response combinations. Refining the displacement further improves every combination, achieving the best performance in all 12 comparisons.

The benefit of refinement is particularly clear for DINO. While pretrained displacement degrades both second-order responses, from $72.16\%$ to $70.48\%$ for $\Delta_Z$ and from $72.23\%$ to $71.77\%$ for $\Delta_H$, the refined responses recover and surpass their raw counterparts, reaching $72.66\%$ and $73.71\%$, respectively. These results are consistent with the relatively modest gain of DSProbe over MVProbe under the fixed pretrained reference, particularly for DINO (Table~\ref{tab:main}): the improvement comes primarily from the first-order response, while the second-order displacements provide less gain. Once the reference is refined, both second-order responses also surpass their raw counterparts, yielding more consistent accuracy.

\begin{table}[t]
\caption{Single-branch source-class membership accuracy (\%) of raw, DSProbe, and
R-DSProbe responses on Model Jungle discriminative models. Values are mean $\pm$
population standard deviation over five seeds. Best per model within each
response type in bold.}
\label{tab:single_branch_response}
\centering
\vspace{3pt}
\resizebox{\linewidth}{!}{%
\begin{tabular}{lccc|ccc|ccc}
\toprule
& \multicolumn{3}{c|}{\textbf{First-order}}
& \multicolumn{3}{c|}{\textbf{Left Gram}}
& \multicolumn{3}{c}{\textbf{Right Gram}} \\
\cmidrule(lr){2-4}
\cmidrule(lr){5-7}
\cmidrule(lr){8-10}

\textbf{Model}
& $W$
& $\Delta_W$
& $\widetilde{\Delta}_W$
& $WW^\top$
& $\Delta_Z$
& $\widetilde{\Delta}_Z$
& $W^\top W$
& $\Delta_H$
& $\widetilde{\Delta}_H$ \\
\midrule

ResNet
& $89.18 \pm 0.30$
& $91.10 \pm 0.28$
& $\mathbf{91.78 \pm 0.28}$
& $90.69 \pm 0.54$
& $93.82 \pm 0.24$
& $\mathbf{93.96 \pm 0.31}$
& $87.47 \pm 0.30$
& $94.45 \pm 0.22$
& $\mathbf{94.85 \pm 0.23}$ \\

SupViT
& $87.78 \pm 0.22$
& $91.11 \pm 0.36$
& $\mathbf{92.04 \pm 0.27}$
& $89.30 \pm 0.31$
& $92.35 \pm 0.46$
& $\mathbf{93.10 \pm 0.19}$
& $90.70 \pm 0.13$
& $93.65 \pm 0.24$
& $\mathbf{94.07 \pm 0.14}$ \\

MAE
& $76.73 \pm 0.10$
& $77.35 \pm 0.21$
& $\mathbf{78.42 \pm 0.29}$
& $77.84 \pm 0.12$
& $78.85 \pm 0.26$
& $\mathbf{80.26 \pm 0.30}$
& $78.26 \pm 0.42$
& $79.28 \pm 0.24$
& $\mathbf{80.38 \pm 0.14}$ \\

DINO
& $68.29 \pm 0.26$
& $69.47 \pm 0.55$
& $\mathbf{71.57 \pm 0.31}$
& $72.16 \pm 0.12$
& $70.48 \pm 0.32$
& $\mathbf{72.66 \pm 0.16}$
& $72.23 \pm 0.18$
& $71.77 \pm 0.20$
& $\mathbf{73.71 \pm 0.21}$ \\

\bottomrule
\end{tabular}%
}
\end{table}

\section{Is the Reference Just a Learnable Response Offset?}
\label{app:offset}

Because the probe is linear, subtracting a reference from the weight is equivalent
to adding a constant, model-independent offset to the response. For a fixed probe
$U$ and a reference $W_{\mathrm{ref}}$ shared by every leaf of a tree,
\begin{equation}
(W_l - W_{\mathrm{ref}})\,U = W_l U + B,
\qquad B = -\,W_{\mathrm{ref}}\,U,
\end{equation}
and likewise $(W_l^\top W_l - W_{\mathrm{ref}}^\top W_{\mathrm{ref}})\,U = W_l^\top
W_l\,U + B'$ for the second-order branches. DSProbe is therefore MVProbe with a
per-branch additive offset applied to the response before standardization and the
projection, and R-DSProbe is the case where that offset is produced by a single
learned matrix $W_*$ shared across the four branches. This raises a natural
concern: if the reference merely acts as a response offset, its benefit might come
from the added capacity of that offset rather than from what is subtracted.

This concern is sharpened by the fact that a free offset is both more expressive
and cheaper than ours. A per-branch offset $B \in \R^{m\times r}$ learned
independently for each of the four branches strictly contains our parameterization,
in which all four offsets are generated by the single matrix $W_*$; it is also
smaller, at $0.39$M parameters against $0.59$M for $W_*$ on a $768\times768$ layer.
If such a free offset matched R-DSProbe, the contribution would reduce to choosing a
good initialization for a bias term. We therefore measure it directly, replacing
$W_*$ with a free per-branch offset $B$ added to the response and trained end to
end, under two initializations---zero and the centering point
$B = -W_{\mathrm{mean}}U$---and compare against R-DSProbe (Table~\ref{tab:offset}).

Two findings follow. First, a free offset initialized at zero adds $0.39$M
trainable parameters yet recovers none of the gain, showing no consistent improvement over MVProbe across the four trees: capacity in the response is not what the reference supplies. Second, a
free offset initialized at the centering point starts exactly where
$W - W_{\mathrm{mean}}$ starts, yet the results show no consistent gains over mean centering alone and on SupViT and MAE drifts below it. With both offsets learned
freely under the same initialization, the weight-space reference $W_*$ still
outperforms the free response-space offset on every architecture (e.g., MAE
$85.03$ vs.\ $83.28$, SupViT $95.95$ vs.\ $94.92$), isolating the
parameterization---not the training constraint---as the operative difference.

Expressiveness is therefore not the operative variable; what matters is where the
offset lives. Our reference is parameterized in weight space, so the quantity
actually subtracted, $-W_*U$, is recomputed from the current probes at every step
and remains the response of a single reference as $U$ is learned---the four branches
move together because they are generated by one matrix. A free offset instead lives
in response space: it is untied from $U$, and the two drift apart as training
proceeds. The coupling that makes our parameterization \emph{less} expressive is
a plausible source of its predictive advantage. We accordingly state the claim of
\Secref{sec:discussion} as: the gains come from subtracting a reference defined
in weight space and shared across branches, not from additional capacity in the
probing head.

\begin{table}[t]
\caption{A free learnable offset on the response versus the reference it is meant to
subsume, under the four-branch architecture. $B$ is a per-branch offset added to the
response before standardization and projection; R-DSProbe produces the offset from a
single shared $W_*$. All learned variants are uncapped. Accuracy (\%), mean over
five seeds; best per column in bold.}
\label{tab:offset}
\centering
\resizebox{\linewidth}{!}{%
\begin{tabular}{llcccc}
\toprule
\textbf{Probe input} & \textbf{Method} & \textbf{ResNet} & \textbf{SupViT} & \textbf{MAE} & \textbf{DINO}\\
\midrule
$W_lU$ & MVProbe
  & $91.09$ & $92.41$ & $81.24$ & $77.88$\\
\addlinespace[2pt]
$W_lU + B$ (init $0$) & Free offset
  & $91.11$ & $92.59$ & $78.49$ & $77.70$\\
$W_lU + B$ (init $-W_{\mathrm{mean}}U$) & Free offset
  & $96.70$ & $94.92$ & $83.28$ & $78.50$\\
\addlinespace[2pt]
$W_l - W_*$ & \textbf{R-DSProbe (ours)}
  & $\mathbf{97.32}$ & $\mathbf{95.95}$ & $\mathbf{85.03}$ & $\mathbf{79.01}$\\
\bottomrule
\end{tabular}%
}
\end{table}

\section{Robustness to Layer Selection}
\label{app:robustness}

Probing-based methods predict from the weights of a single layer, so they must
choose which layer to probe. Because a network's representations are heterogeneous
across depth, accuracy varies substantially from layer to layer, and the optimal
layer differs across architectures and tasks and is not known in advance. Finding
it requires training and validating a metanetwork at several layers, and a method
that is sensitive to this choice can degrade sharply when the wrong layer is
selected. A probing method should therefore be robust to layer selection.
\citet{mvprobe} likewise compare across informative intermediate layers on this
basis, using the layer-index range $\{20,\dots,73\}$. That index, however, follows
the repository's lexicographic enumeration rather than the forward pass
(Appendix~\ref{app:layer_ordering}), so we instead reorder the layers into forward
order and define the informative range there. In forward order the accuracy forms a
unimodal profile that peaks at the central depth and decreases toward both ends
(\Figref{fig:layerwise}); we therefore compare the two methods over the central
$70\%$ of the forward order, excluding the low-information ends (the early
embedding and shallow blocks, and the final classifier).

Table~\ref{tab:robustness} compares R-DSProbe and MVProbe over the central $70\%$ of
the forward order. R-DSProbe outperforms MVProbe at every layer in this range on
ResNet, MAE, and DINO (a $100\%$ win rate), and at $82.7\%$ of layers on SupViT,
for a mean win rate of $95.7\%$. Moreover, both the median and the minimum accuracy
over this range are higher than MVProbe's on all four architectures. This means
that R-DSProbe retains higher performance even when the optimal layer is not selected
exactly, indicating robustness to layer selection.

\begin{table}[t]
\caption{Robustness to layer selection over the central $70\%$ of the forward
order (\Figref{fig:layer_ordering}). Win rate is the fraction of layers where
R-DSProbe exceeds MVProbe; median and min report the typical and worst-case accuracy
(\%) across those layers.}
\label{tab:robustness}
\centering
\begin{tabular}{lccc}
\toprule
& & \textbf{Median (\%)} & \textbf{Min (\%)}\\
\textbf{Model} & \textbf{Win rate (\%)} & \textbf{R-DSProbe / MVProbe} & \textbf{R-DSProbe / MVProbe}\\
\midrule
ResNet & $100.0$ & $87.7$ / $77.2$ & $58.9$ / $55.7$\\
SupViT & $82.7$  & $86.3$ / $78.6$ & $67.2$ / $64.1$\\
MAE    & $100.0$ & $73.9$ / $65.3$ & $57.4$ / $55.4$\\
DINO   & $100.0$ & $71.2$ / $65.9$ & $59.9$ / $54.6$\\
\midrule
Average & $95.7$ & $79.8$ / $71.7$ & $60.9$ / $57.4$\\
\bottomrule
\end{tabular}
\end{table}

\section{Effect of Reference Initialization and Training Details}
\label{app:reference_init}
We further examine how the initialization and training configuration of the learned reference $W_*$ affect R-DSProbe.
In Table~\ref{tab:reference_init}, we first compare initialization from the mean training weight $W_{\mathrm{mean}}$, the pretrained weight $W_{\mathrm{pre}}$, a random matrix $W_{\mathrm{rand}}$, and a random matrix with matched norm to $W_{\mathrm{pre}}$.
The initialization of the reference has a substantial effect on downstream accuracy.
Initializing from either $W_{\mathrm{mean}}$ or $W_{\mathrm{pre}}$ consistently outperforms random initialization,
indicating that a reference close to the shared weight structure provides a substantially better starting point for refinement.
In particular, each leaf admits the decomposition $W_l=W_{\mathrm{pre}} + C +\widetilde{\Delta}_W$. Taking the expectation over leaves gives
\begin{equation}
\mathbb{E}_l[W_l]
=
W_{\mathrm{mean}}
=
W_{\mathrm{pre}}+C + \mathbb{E}_l[\widetilde{\Delta}_W].
\end{equation}
Hence, $W_{\mathrm{mean}}$ decomposes into the pretrained origin $W_{\mathrm{pre}}$, the additional shared component $C$, and the mean refined displacement $\mathbb{E}[\widetilde{\Delta}_W]$, making it a natural alternative initialization for $W_*$. Empirically, $W_{\mathrm{mean}}$ initialization outperforms $W_{\mathrm{pre}}$ on SupViT, MAE, and DINO, so we use it by default.
Importantly, Table~\ref{tab:reference} shows that $W_{\mathrm{mean}}$ is not consistently a better fixed reference than $W_{\mathrm{pre}}$, while it provides a better initialization for learning $W_*$.

One possible explanation is that centering by $W_{\mathrm{mean}}$ removes the mean displacement component $\mathbb{E}_l[\widetilde{\Delta}_W]$ in addition to the shared correction $C$, which may still contain predictive information.
Thus, $W_{\mathrm{mean}}$ can provide a favorable starting point for learning the shared reference, while being overly aggressive when used directly as a fixed reference.
Consistent with this interpretation, Table~\ref{tab:reference_distance} shows that the learned reference moves closer to \(W_{\mathrm{pre}}\) only for ResNet, the only architecture where \(W_{\mathrm{pre}}\) is also a better initialization than \(W_{\mathrm{mean}}\).
Here, $d_{\mathrm{pre}}(W')$ denotes relative Frobenius distance from the pretrained weight of an arbitrary weight matrix $W'$:
\begin{equation}
    d_{\mathrm{pre}}(W')={\|W'-W_{\mathrm{pre}}\|_F\over\|W_{\mathrm{pre}}\|_F}\times 100
\end{equation}
In contrast, it moves farther away for SupViT, MAE, and DINO, suggesting that learning \(W_*\) adapts the reference toward a tree-specific shared component.

Having established $W_{\mathrm{mean}}$ as a suitable initialization, we next examine the optimization of the reference itself. We observed that $W_*$ can undergo overly large updates during training, potentially destabilizing the joint optimization of the reference and the probing network. To stabilize this refinement, we cap the Frobenius norm of each post-Adam reference update at \(\rho\|W_{\mathrm{mean}}\|_F\). This limits individual update steps without constraining the cumulative distance from the initialization. As shown in Table~\ref{tab:reference_init}, this simple constraint improves performance across all four architectures.
We further sweep the cap value in Table~\ref{tab:reference_cap}. A cap of $10^{-3}$ provides the most consistent performance across architectures, achieving the best accuracy on ResNet, SupViT, MAE, and DINO. We therefore use $10^{-3}$ as the default cap throughout our experiments.

\begin{table}[t]
\centering
\caption{
Ablation on reference initialization and training configuration of R-DSProbe.
Values are reported as mean $\pm$ population standard deviation over five seeds.
Best results are shown in bold and second-best results are underlined.
}
\label{tab:reference_init}
\resizebox{\columnwidth}{!}{%
\begin{tabular}{l|c|cccc|c}
\toprule
\multirow{2}{*}{\textbf{Model}}
& \multirow{2}{*}{\textbf{MVProbe}}
& \multicolumn{5}{c}{Reference initialization of \textbf{R-DSProbe}} \\
\cmidrule(lr){3-7}
&
& \textbf{$W_{\mathrm{mean}}$}
& \textbf{$W_{\mathrm{pre}}$}
& \textbf{$W_{\mathrm{rand}}$}
& \textbf{$W_{\mathrm{rand}}$ + Norm-matched}
& \textbf{$W_{\mathrm{mean}}$ + Cap (ours)} \\
\midrule

ResNet
& $91.089 \pm 0.329$
& $97.315 \pm 0.054$
& $\mathbf{97.402 \pm 0.183}$
& $89.326 \pm 0.741$
& $92.771 \pm 1.914$
& $\underline {97.330 \pm 0.051}$ \\

SupViT
& $92.413 \pm 0.398$
& $\underline{95.952 \pm 0.072}$
& $95.917 \pm 0.177$
& $92.094 \pm 0.573$
& $91.632 \pm 0.149$
& $\mathbf{96.511 \pm 0.187}$ \\

MAE
& $81.238 \pm 0.195$
& $\underline{85.028 \pm 0.267}$
& $84.713 \pm 0.251$
& $82.799 \pm 0.350$
& $83.002 \pm 0.314$
& $\mathbf{85.243 \pm 0.265}$ \\

DINO
& $77.879 \pm 0.144$
& $\underline{79.011 \pm 0.227}$
& $78.910 \pm 0.224$
& $75.169 \pm 0.244$
& $76.959 \pm 0.312$
& $\mathbf{79.113 \pm 0.132}$ \\

\bottomrule
\end{tabular}%
}
\end{table}

\begin{table}[t]
\centering
\caption{
Relative Frobenius distance of the initial $W_{\mathrm{mean}}$ and the learned reference $W_*$ from the pretrained weight $W_{\mathrm{pre}}$.
Values for $W_*$ are reported as mean $\pm$ population standard deviation over five seeds.
}
\label{tab:reference_distance}
\begin{tabular}{lcc|c}
\toprule
\textbf{Model}
& \textbf{$d_{\mathrm{pre}}(W_{\mathrm{mean}})$}
& \textbf{$d_{\mathrm{pre}}(W_{*})$}
& \textbf{Change in $d_{\mathrm{pre}}$} \\
\midrule

ResNet
& $1.591\%$
& $1.489 \pm 0.081\%$
& $\mathbf{-0.102\%}$ \\

SupViT
& $1.813\%$
& $3.750 \pm 0.241\%$
& $\mathbf{+1.937\%}$ \\

MAE
& $2.069\%$
& $2.416 \pm 0.205\%$
& $\mathbf{+0.347\%}$ \\

DINO
& $3.971\%$
& $4.363 \pm 0.034\%$
& $\mathbf{+0.392\%}$ \\

\bottomrule
\end{tabular}
\end{table}

\begin{table}[t]
\centering
\caption{
Ablation on the reference update cap.
Values are reported as mean $\pm$ population standard deviation over five seeds.
Best results are shown in bold.
}
\label{tab:reference_cap}
\resizebox{\columnwidth}{!}{%
\begin{tabular}{lcccc}
\toprule
\textbf{Cap}
& \textbf{ResNet}
& \textbf{SupViT}
& \textbf{MAE}
& \textbf{DINO} \\
\midrule

Uncap (0)
& $97.315 \pm 0.054$
& $95.952 \pm 0.072$
& $85.028 \pm 0.267$
& $79.011 \pm 0.227$ \\

$1\times10^{-4}$
& $97.294 \pm 0.058$
& $96.076 \pm 0.159$
& $85.240 \pm 0.216$
& $78.725 \pm 0.292$ \\

$1\times10^{-3}$
& $\mathbf{97.330 \pm 0.051}$
& $\mathbf{96.511 \pm 0.187}$
& $\mathbf{85.243 \pm 0.265}$
& $\mathbf{79.113 \pm 0.132}$ \\

$1\times10^{-2}$
& $97.315 \pm 0.054$
& $95.952 \pm 0.072$
& $85.028 \pm 0.267$
& $79.074 \pm 0.192$ \\

$1\times10^{-1}$
& $97.315 \pm 0.054$
& $95.952 \pm 0.072$
& $85.028 \pm 0.267$
& $79.011 \pm 0.227$ \\

\bottomrule
\end{tabular}%
}
\end{table}

\section{Tree Membership from Weights}
\label{app:tree_membership}

This section gives the tables supporting \Secref{sec:noroot}.
Table~\ref{tab:tree_separation} reports how well the three ViT-B trees are
separated in weight space at each tree's probing layer, and
Table~\ref{tab:tree_contamination} reports the accuracy of R-DSProbe when a
fraction $p$ of each tree's training leaves is replaced by the nearest leaves
of other trees.

\begin{table}[t]
\caption{Separation of the trees in weight space (Frobenius distance): largest distance of a leaf to its own tree mean / smallest distance to another tree's mean (ratio).}
\label{tab:tree_separation}
\centering
\begin{tabular}{lccc}
\toprule
\textbf{Layer} & \textbf{SupViT} & \textbf{MAE} & \textbf{DINO}\\
\midrule
L53 & $15.9 / 56.1\ (\times3.5)$ & $21.2 / 36.0\ (\times1.7)$ & $15.6 / 36.3\ (\times2.3)$\\
L65 & $16.4 / 59.0\ (\times3.6)$ & $20.6 / 39.1\ (\times1.9)$ & $13.5 / 39.2\ (\times2.9)$\\
L47 & $15.4 / 55.1\ (\times3.6)$ & $20.2 / 33.9\ (\times1.7)$ & $17.5 / 34.1\ (\times1.9)$\\
\bottomrule
\end{tabular}
\end{table}

\begin{table}[t]
\caption{R-DSProbe accuracy (\%) when a fraction $p$ of each tree's training leaves is replaced by the nearest leaves of other trees (default cap; $p=0\%$ is unmodified). Mean $\pm$ std over five seeds.}
\label{tab:tree_contamination}
\centering
\begin{tabular}{lccc}
\toprule
\textbf{Tree (layer)} & $p=0\%$ & $p=5\%$ & $p=10\%$\\
\midrule
SupViT (L53) & $96.27 \pm 0.09$ & $93.98 \pm 0.19$ & $92.00 \pm 0.24$\\
MAE (L65) & $85.36 \pm 0.23$ & $82.77 \pm 0.29$ & $80.08 \pm 0.17$\\
DINO (L47) & $79.07 \pm 0.18$ & $76.53 \pm 0.19$ & $75.48 \pm 0.27$\\
\bottomrule
\end{tabular}
\end{table}

\section{Statistical Significance of the SD-LoRA Results}
\label{app:significance}

To test whether the improvements of R-DSProbe over MVProbe on Stable Diffusion LoRA models are statistically significant, we run paired t-tests over five matched random seeds, with the same setup and evaluation protocol as Table~\ref{tab:main}. Table~\ref{tab:significance} reports the results. On SD-200, R-DSProbe is comparable to MVProbe: the differences are small relative to the seed-to-seed variation and not statistically significant, which may partly reflect the already strong performance of MVProbe ($99.80\%$ ID and $95.25\%$ held-out-class accuracy), leaving limited room for further improvement. On SD-1k, the improvements are larger and statistically significant: R-DSProbe improves ID accuracy by $1.042$ points ($p=0.0021$) and held-out-class accuracy by $1.397$ points ($p=0.0018$), and both remain significant after Holm correction for the four comparisons (adjusted $p=0.0072$).

\begin{table}[t]
\caption{Statistical significance of R-DSProbe over MVProbe on Stable Diffusion LoRA, with the same setup as Table~\ref{tab:main}. Accuracy (\%), mean $\pm$ population standard deviation over five seeds. The $p$-values are from paired t-tests.}
\label{tab:significance}
\centering
\begin{tabular}{lcccc}
\toprule
\textbf{Evaluation} & \textbf{MVProbe} & \textbf{R-DSProbe} & \textbf{Gain (points)} & \textbf{$p$-value}\\
\midrule
SD-200 ID & $99.800 \pm 0.127$ & $99.840 \pm 0.150$ & $+0.040$ & $0.6213$\\
SD-200 held-out & $95.250 \pm 1.425$ & $95.529 \pm 0.753$ & $+0.279$ & $0.5520$\\
SD-1k ID & $\mathbf{97.996} \pm 0.283$ & $\mathbf{99.038} \pm 0.234$ & $\mathbf{+1.042}$ & $\mathbf{0.0021}$\\
SD-1k held-out & $\mathbf{97.086} \pm 0.430$ & $\mathbf{98.483} \pm 0.240$ & $\mathbf{+1.397}$ & $\mathbf{0.0018}$\\
\bottomrule
\end{tabular}
\end{table}

\section{Layer Ordering}
\label{app:layer_ordering}

We treat each model as a sequence of layer weight matrices, and the order in which
these matrices are enumerated defines the horizontal axis of our layer-wise
analysis. The Model Jungle benchmark~\citep{probex} enumerates the weight matrices
of each model in lexicographic order of their parameter names, and prior probing
work~\citep{probex,mvprobe} together with the main text refers to layers by this
index (e.g., ``layer 59''). This enumeration, however, does not follow the
network's forward pass: because multi-digit block indices are sorted
lexicographically, the enumeration proceeds through blocks
$0, 1, 10, 11, 2, \dots, 9$, so that block $10$ immediately follows block $1$.
The classifier head and the position embedding are placed at indices $0$ and $1$,
putting the layers that come last and first in the forward pass at the front of the
sequence.

\begin{figure}[t]
\centering
\includegraphics[width=\textwidth]{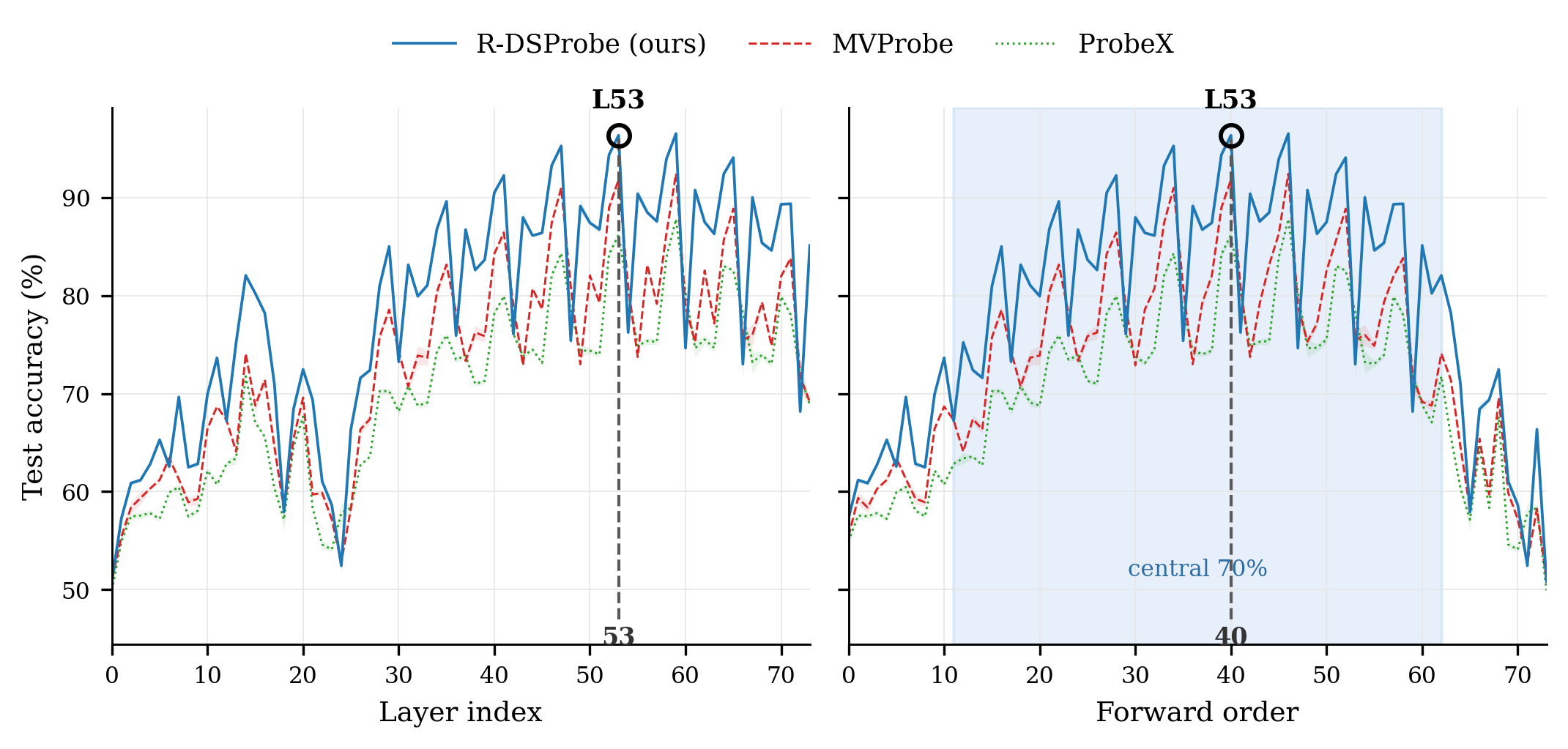}
\caption{Effect of layer ordering on SupViT. Test accuracy (\%) of the three
methods plotted against the repository enumeration index (left) and the recovered
forward-pass position (right). The same weight matrix (L53) moves from index $53$
to forward position $40$.}
\label{fig:layer_ordering}
\end{figure}

Each index still identifies a specific weight matrix; ``layer 53,'' for instance,
denotes the attention-output weight of block~6 and is used with this meaning
throughout. What the index does not reflect is the network's depth. For every
layer-wise analysis (\Figref{fig:layerwise}, Appendix~\ref{app:robustness}) we
therefore keep the matrix that each index denotes but reorder the matrices by their
forward-pass position, recovered from the parameter names (block/stage number and
within-block position), and apply the same forward order to all three methods.
\Figref{fig:layer_ordering} illustrates the effect on SupViT: the same weight
matrix (L53) sits at position $53$ under the repository index but at position $40$
in forward order, and reordering turns the jagged repository-index profile into a
coherent depth trend.

\section{Proof of Higher-Order Separation Theorem}
\label{app:proof_higher_order}

We provide the proof of
Theorem~\ref{thm:higher_order_app}.
Recall the alternating probing hierarchy
\[
\Phi_{2q}(X;U)
=
(X^\top X)^q U,
\qquad
\Phi_{2q+1}(X;U)
=
X(X^\top X)^q U,
\qquad q\ge 0.
\]

The proof proceeds in two steps.
We first establish sufficiency by explicitly constructing two matrices
whose probing responses coincide up to order \(k-1\) and separate
at order \(k\).
We then prove necessity using the growth of the corresponding
block-Krylov reachable subspaces.

\subsection{Preliminaries}

We begin by converting the alternating multiplication by
\(X\) and \(X^\top\) into powers of a single symmetric operator.

\begin{lemma}[Symmetric lifting]
\label{lem:symmetric_lifting}
For \(X\in\mathbb{R}^{m\times n}\), define
\[
B_X
:=
\begin{bmatrix}
0 & X\\
X^\top & 0
\end{bmatrix}
\in
\mathbb{R}^{(m+n)\times(m+n)},
\qquad
\bar U
:=
\begin{bmatrix}
0\\
U
\end{bmatrix}.
\]
Then, for every \(q\ge0\),
\[
B_X^{2q}\bar U
=
\begin{bmatrix}
0\\
\Phi_{2q}(X;U)
\end{bmatrix},
\qquad
B_X^{2q+1}\bar U
=
\begin{bmatrix}
\Phi_{2q+1}(X;U)\\
0
\end{bmatrix}.
\]
\end{lemma}

\begin{proof}
Direct multiplication gives
\[
B_X^2
=
\begin{bmatrix}
XX^\top & 0\\
0 & X^\top X
\end{bmatrix}.
\]
Hence
\[
B_X^{2q}
=
\begin{bmatrix}
(XX^\top)^q & 0\\
0 & (X^\top X)^q
\end{bmatrix},
\]
which yields
\[
B_X^{2q}\bar U
=
\begin{bmatrix}
0\\
(X^\top X)^q U
\end{bmatrix}.
\]
Multiplying once more by \(B_X\) gives
\[
B_X^{2q+1}\bar U
=
\begin{bmatrix}
X(X^\top X)^qU\\
0
\end{bmatrix}.
\]
\end{proof}

We next record a perturbation identity that will allow us to
characterize the first order at which two probing sequences differ.

\begin{lemma}[Perturbation recurrence]
\label{lem:perturbation_recurrence}
Let
\[
B_1:=B_{X_1},
\qquad
B_2:=B_{X_2},
\qquad
E:=B_2-B_1,
\]
and define
\[
D_j
:=
B_2^j\bar U-B_1^j\bar U.
\]
Then
\[
D_{j+1}
=
B_2D_j
+
EB_1^j\bar U.
\]
Consequently, if
\[
D_j=0,
\qquad
0\le j<k,
\]
then
\[
EB_1^j\bar U=0,
\qquad
0\le j\le k-2,
\]
and
\[
D_k
=
EB_1^{k-1}\bar U.
\]
\end{lemma}

\begin{proof}
Using \(B_2=B_1+E\),
\begin{align}
D_{j+1}
&=
B_2^{j+1}\bar U-B_1^{j+1}\bar U
\\
&=
B_2
\left(
B_2^j\bar U-B_1^j\bar U
\right)
+
(B_2-B_1)B_1^j\bar U
\\
&=
B_2D_j+EB_1^j\bar U.
\end{align}
If \(D_j=D_{j+1}=0\), the recurrence implies
\(EB_1^j\bar U=0\).
Applying this for \(j=0,\ldots,k-2\) gives the first claim,
while setting \(D_{k-1}=0\) gives
\[
D_k
=
EB_1^{k-1}\bar U.
\]
\end{proof}

\subsection{Sufficiency}
\label{app:sufficiency}

Suppose
\[
n-r
\ge
\left\lfloor\frac{k}{2}\right\rfloor,
\qquad
m
\ge
\left\lceil\frac{k}{2}\right\rceil.
\]
Let
\[
s
:=
\left\lfloor\frac{k}{2}\right\rfloor,
\qquad
\ell
:=
\left\lceil\frac{k}{2}\right\rceil.
\]

\paragraph{Step 1: Construct an alternating orthonormal chain.}

Choose a unit vector
\[
p_0\in\operatorname{range}(U).
\]
Then
\[
p_0^\top U\neq 0.
\]
Since
\[
\dim\ker(U^\top)
=
n-r
\ge s,
\]
there exist orthonormal vectors
\[
p_1,\ldots,p_s
\in
\ker(U^\top),
\]
so that
\[
p_i^\top U=0,
\qquad
i=1,\ldots,s.
\]
Similarly, since \(m\ge\ell\), choose orthonormal vectors
\[
q_1,\ldots,q_\ell
\in\mathbb{R}^m.
\]

Lift these vectors to \(\mathbb{R}^{m+n}\) by defining
\[
z_{2i}
:=
\begin{bmatrix}
0\\
p_i
\end{bmatrix},
\qquad
z_{2i-1}
:=
\begin{bmatrix}
q_i\\
0
\end{bmatrix}.
\]
Then
\[
z_0,z_1,\ldots,z_k
\]
form an orthonormal sequence alternating between the input and
output spaces.

\paragraph{Step 2: Construct \(X_1\) and \(X_2\).}

For \(j=1,\ldots,k\), define
\[
M_j
:=
\begin{cases}
q_{i+1}p_i^\top,
&
j=2i+1,
\\[2mm]
q_i p_i^\top,
&
j=2i.
\end{cases}
\]
Set
\[
X_1
:=
\sum_{j=1}^{k-1}M_j,
\qquad
X_2
:=
\sum_{j=1}^{k}M_j.
\]
Hence
\[
X_2-X_1=M_k,
\]
which is rank one, and therefore \(X_1\neq X_2\).

\paragraph{Step 2 (continued): Lifted operators.}
Recall the lift $A\mapsto B_A:=\begin{bmatrix}0&A\\ A^\top&0\end{bmatrix}$ of
Lemma~\ref{lem:symmetric_lifting}, which is linear in $A$. Hence
\[
B_{X_1}
=B_{\sum_{j=1}^{k-1}M_j}
=\begin{bmatrix}0&\sum_{j=1}^{k-1}M_j\\[2pt]\sum_{j=1}^{k-1}M_j^\top&0\end{bmatrix}
=\sum_{j=1}^{k-1}B_{M_j},
\qquad
B_{X_2}=\sum_{j=1}^{k}B_{M_j}.
\]
For each $j$, the term $M_j$ couples the consecutive chain vectors
$z_{j-1}$ and $z_j$, which lie in complementary blocks; writing
$z_{j-1}=\begin{bmatrix}q\\ 0\end{bmatrix}$ and
$z_{j}=\begin{bmatrix}0\\ p\end{bmatrix}$ with $M_j=qp^\top$,
\[
z_{j-1}z_j^\top=\begin{bmatrix}0&qp^\top\\ 0&0\end{bmatrix},
\qquad
z_{j}z_{j-1}^\top=\begin{bmatrix}0&0\\ pq^\top&0\end{bmatrix},
\]
so that
\[
B_{M_j}
=\begin{bmatrix}0&M_j\\ M_j^\top&0\end{bmatrix}
=z_{j}z_{j-1}^\top+z_{j-1}z_j^\top.
\]
Summing over $j$,
\[
B_1:=B_{X_1}=\sum_{j=1}^{k-1}\bigl(z_jz_{j-1}^\top+z_{j-1}z_j^\top\bigr),
\qquad
B_2:=B_{X_2}=B_1+z_kz_{k-1}^\top+z_{k-1}z_k^\top,
\]
and therefore the perturbation is the single terminal coupling
\[
E:=B_2-B_1=z_kz_{k-1}^\top+z_{k-1}z_k^\top,
\]
which acts only on $\operatorname{span}\{z_{k-1},z_k\}$.

\paragraph{Step 3: Track the reachable directions.}

By construction,
\[
B_1z_0=z_1,
\]
and, for \(1\le j\le k-2\),
\[
B_1z_j
=
z_{j-1}+z_{j+1}.
\]
It follows inductively that
\[
B_1^t z_0
=
z_t+v_t,
\qquad
v_t
\in
\operatorname{span}
\{z_0,\ldots,z_{t-1}\},
\qquad
0\le t\le k-1.
\]
In particular, the newly reached direction \(z_t\) appears
with nonzero coefficient at every step.

Let
\[
c
:=
p_0^\top U
\in\mathbb{R}^{1\times r},
\qquad
c\neq0.
\]
Since \(p_i^\top U=0\) for \(i\ge1\), we may write
\[
\bar U
=
z_0c+\bar U_\perp,
\]
where every column of \(\bar U_\perp\) is orthogonal to
\[
\operatorname{span}\{z_0,\ldots,z_k\}.
\]
Since both \(B_1\) and \(E\) act only on this span,
\[
B_1\bar U_\perp=0,
\qquad
E\bar U_\perp=0.
\]

\paragraph{Step 4: All lower-order responses coincide.}

For \(0\le t\le k-2\),
\[
B_1^t\bar U
\in
\operatorname{span}
\{z_0,\ldots,z_{k-2}\}.
\]
On the other hand, \(E\) acts only through the final edge
between \(z_{k-1}\) and \(z_k\). Therefore,
\[
EB_1^t\bar U
=
0,
\qquad
0\le t\le k-2.
\]
Lemma~\ref{lem:perturbation_recurrence} then gives
\[
B_2^j\bar U
=
B_1^j\bar U,
\qquad
0\le j<k.
\]
By Lemma~\ref{lem:symmetric_lifting},
\[
\Phi_j(X_1;U)
=
\Phi_j(X_2;U),
\qquad
1\le j<k.
\]

\paragraph{Step 5: The \(k\)-th response separates.}

At \(t=k-1\),
\[
B_1^{k-1}\bar U
=
z_{k-1}c
+
v,
\qquad
v
\in
\operatorname{span}
\{z_0,\ldots,z_{k-2}\}.
\]
Therefore,
\[
EB_1^{k-1}\bar U
=
z_kc
\neq0.
\]
By Lemma~\ref{lem:perturbation_recurrence},
\[
B_2^k\bar U-B_1^k\bar U
=
z_kc
\neq0.
\]
Applying Lemma~\ref{lem:symmetric_lifting} yields
\[
\Phi_k(X_1;U)
\neq
\Phi_k(X_2;U).
\]
This proves sufficiency.

\subsection{Necessity}
\label{app:necessity}

Suppose there exist distinct
\[
X_1,X_2\in\mathbb{R}^{m\times n}
\]
such that
\[
\Phi_j(X_1;U)
=
\Phi_j(X_2;U),
\qquad
1\le j<k,
\]
while
\[
\Phi_k(X_1;U)
\neq
\Phi_k(X_2;U).
\]

Let
\[
X:=X_1,
\qquad
\Delta:=X_2-X_1,
\qquad
G:=X^\top X.
\]
Since the \(k\)-th responses differ,
\[
\Delta\neq0.
\]

Define the right reachable subspaces
\[
\mathcal{R}_t
:=
\operatorname{col}
\left[
U,\,
GU,\,
\ldots,\,
G^tU
\right]
\subseteq\mathbb{R}^n
\]
and the left reachable subspaces
\[
\mathcal{L}_t
:=
\operatorname{col}
\left[
XU,\,
XGU,\,
\ldots,\,
XG^tU
\right]
\subseteq\mathbb{R}^m,
\]
with
\[
\mathcal{L}_{-1}:=\{0\}.
\]
By construction,
\[
\mathcal{L}_t
=
X\mathcal{R}_t.
\]

\begin{lemma}[No restart after stagnation]
\label{lem:no_restart}
If
\[
\mathcal{R}_t
=
\mathcal{R}_{t-1}
\]
for some \(t\ge1\), then
\[
\mathcal{R}_{t+j}
=
\mathcal{R}_{t-1},
\qquad
\forall j\ge0.
\]
The analogous statement holds for
\(\mathcal{L}_t\).
\end{lemma}

\begin{proof}
If
\[
\mathcal{R}_t
=
\mathcal{R}_{t-1},
\]
then
\[
G^tU\in\mathcal{R}_{t-1},
\]
and hence
\[
G\mathcal{R}_{t-1}
\subseteq
\mathcal{R}_{t-1}.
\]
Thus \(\mathcal{R}_{t-1}\) is \(G\)-invariant, implying that
all subsequent powers \(G^{t+j}U\) remain in the same subspace.
The argument for \(\mathcal{L}_t\) is identical, replacing
\(G=X^\top X\) by \(XX^\top\).
\end{proof}

Since
\[
E
=
\begin{bmatrix}
0 & \Delta\\
\Delta^\top & 0
\end{bmatrix},
\]
Lemmas~\ref{lem:symmetric_lifting}
and~\ref{lem:perturbation_recurrence} translate equality of
the lower-order probing responses into explicit annihilation
conditions. We consider even and odd \(k\) separately.

\paragraph{Case 1: \(k=2q\).}

Equality at all orders below \(2q\) implies
\[
\Delta G^tU
=
0,
\qquad
0\le t\le q-1,
\tag{A.1}
\label{eq:even_right_annihilation}
\]
and
\[
\Delta^\top XG^tU
=
0,
\qquad
0\le t\le q-2.
\tag{A.2}
\label{eq:even_left_annihilation}
\]
The first differing response satisfies
\[
\Delta^\top XG^{q-1}U
\neq0.
\tag{A.3}
\label{eq:even_separation}
\]

\Eqref{eq:even_separation}, together with
\eqref{eq:even_left_annihilation}, implies
\[
\mathcal{L}_{q-1}
\supsetneq
\mathcal{L}_{q-2}.
\]
By Lemma~\ref{lem:no_restart}, the sequence cannot have
stagnated at any earlier stage. Hence
\[
\mathcal{L}_{-1}
\subsetneq
\mathcal{L}_0
\subsetneq
\cdots
\subsetneq
\mathcal{L}_{q-1},
\]
and therefore
\[
\dim(\mathcal{L}_{q-1})
\ge q.
\]
Since
\[
\mathcal{L}_{q-1}
\subseteq\mathbb{R}^m,
\]
we obtain
\[
m\ge q.
\]

Moreover, strict growth of
\(\mathcal{L}_t=X\mathcal{R}_t\)
forces strict growth of the corresponding
\(\mathcal{R}_t\). Since
\[
\dim(\mathcal{R}_0)
=
\operatorname{rank}(U)
=
r,
\]
we have
\[
\dim(\mathcal{R}_{q-1})
\ge
r+q-1.
\]
From \eqref{eq:even_right_annihilation},
\[
\Delta\mathcal{R}_{q-1}
=
\{0\},
\]
so that
\[
\mathcal{R}_{q-1}
\subseteq
\ker(\Delta).
\]
Because \(\Delta\neq0\),
\[
\dim\ker(\Delta)
\le
n-1.
\]
Consequently,
\[
r+q-1
\le
n-1,
\]
which gives
\[
n-r\ge q.
\]
Thus, when \(k=2q\),
\[
n-r\ge\frac{k}{2},
\qquad
m\ge\frac{k}{2}.
\]

\paragraph{Case 2: \(k=2q+1\).}

Equality at all lower orders implies
\[
\Delta G^tU
=
0,
\qquad
0\le t\le q-1,
\tag{A.4}
\label{eq:odd_right_annihilation}
\]
and
\[
\Delta^\top XG^tU
=
0,
\qquad
0\le t\le q-1.
\tag{A.5}
\label{eq:odd_left_annihilation}
\]
The first differing response satisfies
\[
\Delta G^qU
\neq0.
\tag{A.6}
\label{eq:odd_separation}
\]

\Eqref{eq:odd_right_annihilation}
and~\eqref{eq:odd_separation} imply
\[
\mathcal{R}_q
\supsetneq
\mathcal{R}_{q-1}.
\]
By Lemma~\ref{lem:no_restart},
\[
\mathcal{R}_0
\subsetneq
\mathcal{R}_1
\subsetneq
\cdots
\subsetneq
\mathcal{R}_q.
\]
Hence
\[
\dim(\mathcal{R}_q)
\ge
r+q.
\]
Since
\[
\mathcal{R}_q
\subseteq
\mathbb{R}^n,
\]
we obtain
\[
n-r\ge q.
\]

We next show that
\[
\mathcal{L}_{q-1}
\supsetneq
\mathcal{L}_{q-2}.
\]
Suppose, to the contrary, that
\[
\mathcal{L}_{q-1}
=
\mathcal{L}_{q-2}.
\]
Then
\[
XG^{q-1}U
\in
\mathcal{L}_{q-2},
\]
and therefore
\[
G^qU
=
X^\top(XG^{q-1}U)
\in
X^\top\mathcal{L}_{q-2}
\subseteq
\mathcal{R}_{q-1}.
\]
\Eqref{eq:odd_right_annihilation} would then imply
\[
\Delta G^qU=0,
\]
contradicting \eqref{eq:odd_separation}.
Thus
\[
\mathcal{L}_{q-1}
\supsetneq
\mathcal{L}_{q-2}.
\]

Lemma~\ref{lem:no_restart} therefore gives
\[
\dim(\mathcal{L}_{q-1})
\ge q.
\]
Moreover, \eqref{eq:odd_left_annihilation} implies
\[
\Delta^\top\mathcal{L}_{q-1}
=
\{0\},
\]
and hence
\[
\mathcal{L}_{q-1}
\subseteq
\ker(\Delta^\top).
\]
Since \(\Delta\neq0\),
\[
\dim\ker(\Delta^\top)
\le
m-1.
\]
Therefore
\[
q\le m-1,
\]
or equivalently,
\[
m\ge q+1.
\]

Thus, when \(k=2q+1\),
\[
n-r\ge q,
\qquad
m\ge q+1.
\]

Combining the even and odd cases yields
\[
n-r
\ge
\left\lfloor\frac{k}{2}\right\rfloor,
\qquad
m
\ge
\left\lceil\frac{k}{2}\right\rceil.
\]
This proves necessity and completes the proof of
Theorem~\ref{thm:higher_order_app}.
\qed

\subsection{Maximum First-Separation Order}
\label{app:max_separation_order}

Theorem~\ref{thm:higher_order_app} immediately implies
the maximum order at which a pair of matrices can become
distinguishable for the first time.

\begin{corollary}[Maximum first-separation order]
\label{cor:max_separation_order}
Let
\[
U\in\mathbb{R}^{n\times r},
\qquad
\operatorname{rank}(U)=r<n.
\]
The maximum feasible first-separation order is
\[
k_{\max}
=
\min
\left\{
2m,\,
2(n-r)+1
\right\}.
\]
Equivalently, for every integer
\[
2
\le
k
\le
\min\{2m,\,2(n-r)+1\},
\]
there exist
\(X_1\neq X_2\in\mathbb{R}^{m\times n}\)
such that
\[
\Phi_j(X_1;U)
=
\Phi_j(X_2;U),
\qquad
\forall j<k,
\]
while
\[
\Phi_k(X_1;U)
\neq
\Phi_k(X_2;U).
\]
\end{corollary}

\begin{proof}
By Theorem~\ref{thm:higher_order_app}, order \(k\)
is feasible if and only if
\[
\left\lfloor\frac{k}{2}\right\rfloor
\le
n-r,
\qquad
\left\lceil\frac{k}{2}\right\rceil
\le
m.
\]
The largest integer satisfying both inequalities is
\[
k_{\max}
=
\min
\left\{
2m,\,
2(n-r)+1
\right\}.
\]
\end{proof}

\subsection{Special Cases}
\label{app:higher_order_special_cases}

For \(k=2\), the dimension condition reduces to
\[
n-r\ge1,
\qquad
m\ge1.
\]
In this case, the construction admits
\[
X_1=q_1p_0^\top,
\qquad
X_2=q_1(p_0+p_1)^\top,
\]
where
\[
p_0^\top U\neq0,
\qquad
p_1^\top U=0.
\]
Hence
\[
X_1U=X_2U,
\]
while
\[
X_1^\top X_1U
\neq
X_2^\top X_2U.
\]
Thus, the second-order response resolves a collision that is
invisible to the first-order response.

For \(k=3\), the required condition is
\[
n-r\ge1,
\qquad
m\ge2.
\]
Accordingly, there exist \(X_1\neq X_2\) satisfying
\[
X_1U=X_2U,
\]
and
\[
X_1^\top X_1U
=
X_2^\top X_2U,
\]
but
\[
X_1X_1^\top X_1U
\neq
X_2X_2^\top X_2U.
\]
Therefore, the third-order response can strictly refine the
distinguishability induced jointly by the first- and
second-order responses.

\end{document}